\documentclass{article}

\usepackage{arxiv}

\usepackage{pdflscape,rotating,placeins}

\usepackage{mathptmx}
\DeclareMathAlphabet{\mathcal}{OMS}{cmsy}{m}{n}

\RequirePackage{fix-cm}
\usepackage{siunitx}
\usepackage{amsmath,bm}

\usepackage[utf8]{inputenc}
\usepackage[T1]{fontenc}
\usepackage{graphicx}
\usepackage[figurename=Fig.,labelfont=bf,labelsep=period]{caption}
\usepackage{subcaption}
\usepackage{amsfonts}
\usepackage{amssymb}
\usepackage{amsbsy} 

\usepackage[colorlinks=true,citecolor=red,linkcolor=black]{hyperref}

\usepackage{url}            
\usepackage{booktabs}       
\usepackage{nicefrac}       
\usepackage{microtype}      
\usepackage{lipsum}
\usepackage{algorithm,tikz}
\usepackage[]{algpseudocode}
\usepackage{pgfplots}
\usepackage{subcaption}
\usepackage[]{threeparttable}
\usepackage{multirow,bm}

\newcommand{\Dtr}{\mathcal{D}_\mathrm{tr}}
\newcommand{\Dval}{\mathcal{D}_\mathrm{val}}

\newcommand{\Ntr}{{N}_\mathrm{tr}}
\newcommand{\Nval}{{N}_\mathrm{val}}

\newcommand{\um}{\mathbf{u}}

\newcommand{\xm}{\mathbf{x}}

\newcommand{\xii}{\pmb{\xi}}

\newcommand{\thetaa}{\pmb{\theta}}

\newcommand{\epsv}{\varepsilon_\mathrm{val}}

\makeatletter
\newcommand*\dashline{~\rotatebox[origin=c]{90}{\scalebox{0.85}{$\dabar@\dabar@\dabar@$}}~}
\makeatother

\newcommand\Tstrut{\rule{0pt}{2.6ex}}         
\newcommand\Bstrut{\rule[-0.9ex]{0pt}{0pt}}   

\usepackage{tikz}
\usetikzlibrary{matrix,chains,positioning,decorations.pathreplacing,arrows}
\usetikzlibrary{positioning,calc}
\usetikzlibrary{patterns,decorations.pathmorphing,decorations.markings}
\usetikzlibrary{shapes}

\usetikzlibrary{fit}
\tikzstyle{block} = [draw,rectangle,thick,minimum height=2em,minimum width=2em]
\tikzstyle{sum} = [draw,circle,inner sep=0mm,minimum size=2mm]
\tikzstyle{connector} = [->,thick]
\tikzstyle{line} = [thick]
\tikzstyle{branch} = [circle,inner sep=0pt,minimum size=1mm,fill=black,draw=black]
\tikzstyle{guide} = []
\tikzset{%
	every neuron/.style={
		circle,
		draw,
		fill = green!20,
		thick,
		minimum size=1.1cm
	},
	neuron missing/.style={
		draw=none, 
		scale=4,
		fill = none,
		text height=0.333cm,
		execute at begin node=\color{black}$\vdots$
	},
	input neuron/.style={
		circle,
		draw,
		thick,
		fill = yellow!20,
		minimum size=1cm
	},
	output neuron/.style={
		circle,
		draw,
		thick,
		fill = red!20,
		minimum size=1cm
	},
}
\tikzset{%
wind turbine/.pic={
  \tikzset{path/.style={fill, draw=white, ultra thick, line join=round}}
  \path [path] 
    (-.25,0) arc (180:360:.25 and .0625) -- (.0625,3) -- (-.0625,3) -- cycle;
  \foreach \i in {90, 210, 330}{
    \ifcase#1
    \or
      \path [path, shift=(90:3), rotate=\i] 
        (.5,-.1875) arc (270:90:.5 and .1875) arc (90:-90:1.5 and .1875);
    \or
      \path [path, shift=(90:3), rotate=\i] 
        (0,0.125) -- (2,0.125) -- (2,0) -- (0.5,-0.375) -- cycle;
    \or
      \path [path, shift=(90:3), rotate=\i]
        (0,-0.125) arc (180:0:1 and 0.125) -- ++(0,0.125) arc (0:180:1 and 0.25) -- cycle;
    \fi
  }
  \path [path] (0,3) circle [radius=.25];
}} 

\usetikzlibrary{3d,decorations.text,shapes.arrows,positioning,fit,backgrounds}
\tikzset{pics/fake box/.style args={
		#1 with dimensions #2 and #3 and #4}{
		code={
			\draw[gray,ultra thin,fill=#1]  (0,0,0) coordinate(-front-bottom-left) to
			++ (0,#3,0) coordinate(-front-top-right) --++
			(#2,0,0) coordinate(-front-top-right) --++ (0,-#3,0) 
			coordinate(-front-bottom-right) -- cycle;
			\draw[gray,ultra thin,fill=#1] (0,#3,0)  --++ 
			(0,0,#4) coordinate(-back-top-left) --++ (#2,0,0) 
			coordinate(-back-top-right) --++ (0,0,-#4)  -- cycle;
			\draw[gray,ultra thin,fill=#1!80!black] (#2,0,0) --++ (0,0,#4) coordinate(-back-bottom-right)
			--++ (0,#3,0) --++ (0,0,-#4) -- cycle;
			\path[gray,decorate,decoration={text effects along path,text={}}] (#2/2,{2+(#3-2)/2},0) -- (#2/2,0,0);
		}
}}
\tikzset{circle dotted/.style={dash pattern=on .05mm off 2mm,
		line cap=round}}
 
\title{Bi-fidelity Modeling of Uncertain and Partially Unknown Systems Using DeepONets}

\author{
  Subhayan De\\
Smead Aerospace Engineering Sciences\\
University of Colorado\\
Boulder, CO 80303\\
  \texttt{Subhayan.De@colorado.edu} \\
  \And
 Matthew Reynolds \\
  National Renewable Energy Laboratory \\
  Golden, CO 80401\\
  \texttt{Matthew.Reynolds@nrel.gov } \\ 
  \And
 Malik Hassanaly \\
  National Renewable Energy Laboratory \\
  Golden, CO 80401\\
  \texttt{Malik.Hassanaly@nrel.gov } \\
  \And
 Ryan N. King \\
  National Renewable Energy Laboratory \\
  Golden, CO 80401\\
  \texttt{Ryan.King@nrel.gov } \\
  \And
  Alireza Doostan\\
Smead Aerospace Engineering Sciences\\
University of Colorado\\
Boulder, CO 80303\\
  \texttt{doostan@colorado.edu} \\
} 

\begin{document}
\maketitle

\begin{abstract}
Recent advances in modeling large-scale, complex physical systems have shifted research focuses towards data-driven techniques. However, generating datasets by simulating complex systems can require significant computational resources. Similarly, acquiring experimental datasets can prove difficult. For these systems, often computationally inexpensive, but in general inaccurate models, known as the \textit{low-fidelity models}, are available. In this paper, we propose a bi-fidelity modeling approach for complex physical systems, where we model the discrepancy between the true system's response and a \textit{low-fidelity} response in the presence of a small training dataset from the true system's response using a deep operator network (DeepONet), a neural network architecture suitable for approximating nonlinear operators. We apply the approach to systems that have parametric uncertainty and are partially unknown. Three numerical examples are used to show the efficacy of the proposed approach to model uncertain and partially unknown physical systems. 
\end{abstract}


\section{Introduction}

The ubiquitous presence of uncertainty often affects the modeling of a physical system using differential equations. For example, in engineering systems, the sources of uncertainty can include material properties, geometry, or loading conditions. This type of uncertainty is known as \textit{parametric uncertainty} and differs from one realization to another. In contrast, during modeling, some physical phenomena are often ignored or simplified, which leads to the so called \textit{structural uncertainty} \cite{kennedy2001bayesian}. Both types of uncertainties can be present in a real-life engineering system, and should be addressed when developing \textit{digital twins} of the system. Models developed using polynomial chaos expansion \cite{ghanem2003stochastic,xiu2002wiener}, Gaussian process regression \cite{williams2006gaussian,forrester2007multi}, and {other} response surfaces \cite{isukapalli1998stochastic,giunta2006promise,chi2012response} address \textit{parametric uncertainty}, which defines possible realizations of the system. However, the cost of developing these models increases significantly with the dimension of the uncertain variables \cite{doostan2011non}.

Neural networks, commonly used in computer vision \cite{voulodimos2018deep} and pattern recognition problems \cite{basu2010use}, have been used to model physical systems in the presence of uncertainty \cite{tripathy2018deep,de2021uncertainty}. 
To model physical systems, Raissi et al. \cite{raissi2019physics} augmented the standard data discrepancy loss with the residual of the corresponding governing equations. This approach was termed "Physics-informed neural networks." 
Subsequently, a significant amount of work employing this approach to address basic computational mechanics problems came into prominence in the literature, some notable ones being
Karniadakis et al. \cite{karniadakis2021physics}, Cai et al. \cite{cai2021physics}, and Viana and Subramaniyan \cite{viana2021survey}. 
Among these several works address the \textit{parameteric uncertainty} in the system \cite{yang2020physics,yang2021b,zhu2019physics,yang2019adversarial,geneva2020modeling,winovich2019convpde}. 
However, machine learning studies that address \textit{structural uncertainty} in modeling are scarce. Among them, Blasketh et al. \cite{blakseth2021deep} used neural networks combined with governing equations to address modeling and discretization errors. Zhang et al. \cite{zhang2019quantifying} used the physics-informed approach to quantify both the parametric and approximation uncertainty of the neural network. 
Duraisamy and his colleagues further trained neural networks to comprehend the discrepancies between turbulence model prediction and the available data
\cite{parish2016paradigm,singh2017machine,duraisamy2019turbulence}. 




Recently, several neural network architectures capable of approximating mesh-independent solution operators for the governing equations of physical systems have been proposed \cite{lu2019deeponet,li2020neural,li2020multipole,li2020fourier,li2021markov,lu2021learning,li2021physics}, drawing inspiration from the universal approximation theorem for operators in Chen and Chen \cite{chen1995universal}. These architectures overcome two shortcomings of the physics-informed use of neural networks to learn a system's behavior: firstly, they do not depend on the mesh used to generate the training dataset. Secondly, they are solely data-driven and do not require the knowledge of the governing equations. Among these network architectures, the \textit{deep operator network} (\textit{DeepONet}) proposed in Lu et al. \cite{lu2019deeponet,lu2021learning} uses two separate networks --- one with the external source term as input and the other with the coordinates where the system response is sought as input. The former network is known as the \textit{branch network} while the latter is known as the \textit{trunk network}. Outputs from these two networks are then used in a linear regression model to produce the system response. 
Lu et al. \cite{lu2021one} showed how the computational domain can be decomposed into many small subdomains to learn the governing equation with only one solution of the equation. While Cai et al. \cite{cai2021deepm} used DeepONets for a multiphysics problem involving modeling of field variables across multiple scales., Ranade et al. \cite{ranade2021generalized} and Priyadarshini et al. \cite{sharma2021application} employed DeepONets for approximating joint subgrid probability density function in turbulent combustion problems and to model non-equilibrium flow physics respectively. Additionally, Wang et al. \cite{wang2021learning,wang2021long} used the error in satisfying the governing equations to train the DeepONets, with a variational formulation of the governing equations being applied by Goswami et al. \cite{goswami2021physics} to train DeepONets to model brittle fracture. Further comparisons of DeepONet with the iterative neural operator architecture proposed in Li et al. \cite{li2020fourier} were performed in Kovachki et al. \cite{kovachki2021neural} and Lu et al. \cite{lu2021comprehensive}. The approximation errors in the latter neural operator architectures were subsequently discussed in Kovachki et al. \cite{kovachki2021universal}, Lanthaler et al. \cite{lanthaler2021error}, Deng et al. \cite{deng2021convergence}, and Marcati and Schwab \cite{marcati2021exponential}.

The training datasets required to construct the DeepONets for a complex physical system can often be limited due to the computational burden associated with simulating such systems. Similarly, repeating experiments to obtain measurement data can become infeasible. 
In the presence of a small dataset from the true high-fidelity system, we propose herein a modeling strategy with DeepONets that uses a similar but computationally inexpensive model, known as \textit{low-fidelity}, that captures many aspects of the true system's behavior. The discrepancy between the response from the true physical system and the \textit{low-fidelity model} is estimated using DeepONets. The assumption used here is that the modeling of the discrepancy requires only a small training dataset of the true system's response, and as a result, bi-fidelity approaches are often advantageous with scarce training data from the high-fidelity simulations \cite{de2020bi,de2020transfer,de2021neural}. 
In principle, we train a DeepONet that uses the low-fidelity response as the input of the branch network to predict the discrepancy between the low- and high-fidelity data. 
However, we do not incorporate physics loss or error in satisfying the governing equations (as previously employed in physics informed neural networks) to circumvent potential bias incorporation in partially known engineering systems subject to the presence of structural uncertainty.

In this paper, we employ three numerical examples to show the efficacy of the proposed bi-fidelity approach to modeling the response of an uncertain and partially unknown physical system. The first two examples consider a nonlinear oscillator and a heat transfer problem, respectively, where uncertainty is present in some of the parameters of the governing equations. The \textit{low-fidelity} model used in these examples ignores the nonlinearity or some of the uncertain components of the governing equations. 
In the third example,  we model power generated in a wind farm, , with the two and three dimensional simulations respectively serving as \textit{high} and \textit{low-fidelity} model systems. 

The rest of the paper is organized as follows: first, we provide a brief background on DeepONets in Section \ref{sec:back}. Then, we present our proposed approach for modeling the discrepancy between a low-fidelity system response and the true response for uncertain and partially unknown systems in Section \ref{sec:method}. Three numerical examples involving a dynamical system, a heat transfer problem, and a wind farm application are subsequently used in Section \ref{sec:ex} to showcase the efficacy of the proposed approach. Finally, we conclude the paper with a brief discussion on future applications of the proposed approach. 


\section{Background on DeepONets} \label{sec:back}
Let us assume the governing equation of a physical system in a domain $\Omega$ is given by
\begin{equation} \label{eq:gov}
    \mathcal{L}(u)(\xm) = f(\xm), \qquad \xm \in \Omega,
\end{equation}
where $\mathcal{L}$ is a nonlinear differential operator; $\xm$ is the location; $u$ is the response of the system; and $f$ is the external source. 
For a boundary value problem, Dirichlet boundary condition $u(\xm)={u}_D(\xm)$ on $\xm\in\partial \Omega_D$ and/or Neumann boundary condition $\mathcal{L}_N (u)(\xm)={u}_N(\xm)$ on $\xm\in\partial \Omega_N$ are also present, where $\mathcal{L}_N$ is the differential operator for the Neumann boundary condition; $\partial \Omega_D\cup\partial \Omega_N = \partial \Omega$ with the boundary of the domain being $\partial \Omega$; and $\partial \Omega_D\cap\partial \Omega_N = \emptyset$. For an initial value problem, an initial condition $u(\xm_0) = u_0$ is used instead, where $\xm_0$ is the initial instant. {The solution of \eqref{eq:gov} is given by $G(f)(\xm)$, where $G:\mathcal{X}\rightarrow \mathcal{Y}$ denotes the solution operator. For $\mathcal{X}$ and $\mathcal{Y}$, there exist embeddings $\mathcal{X}\hookrightarrow L^2(D)$ and $\mathcal{Y}\hookrightarrow L^2(\Omega)$. } 

\begin{figure}[!htb]

	\centering
	\includegraphics[scale=1]{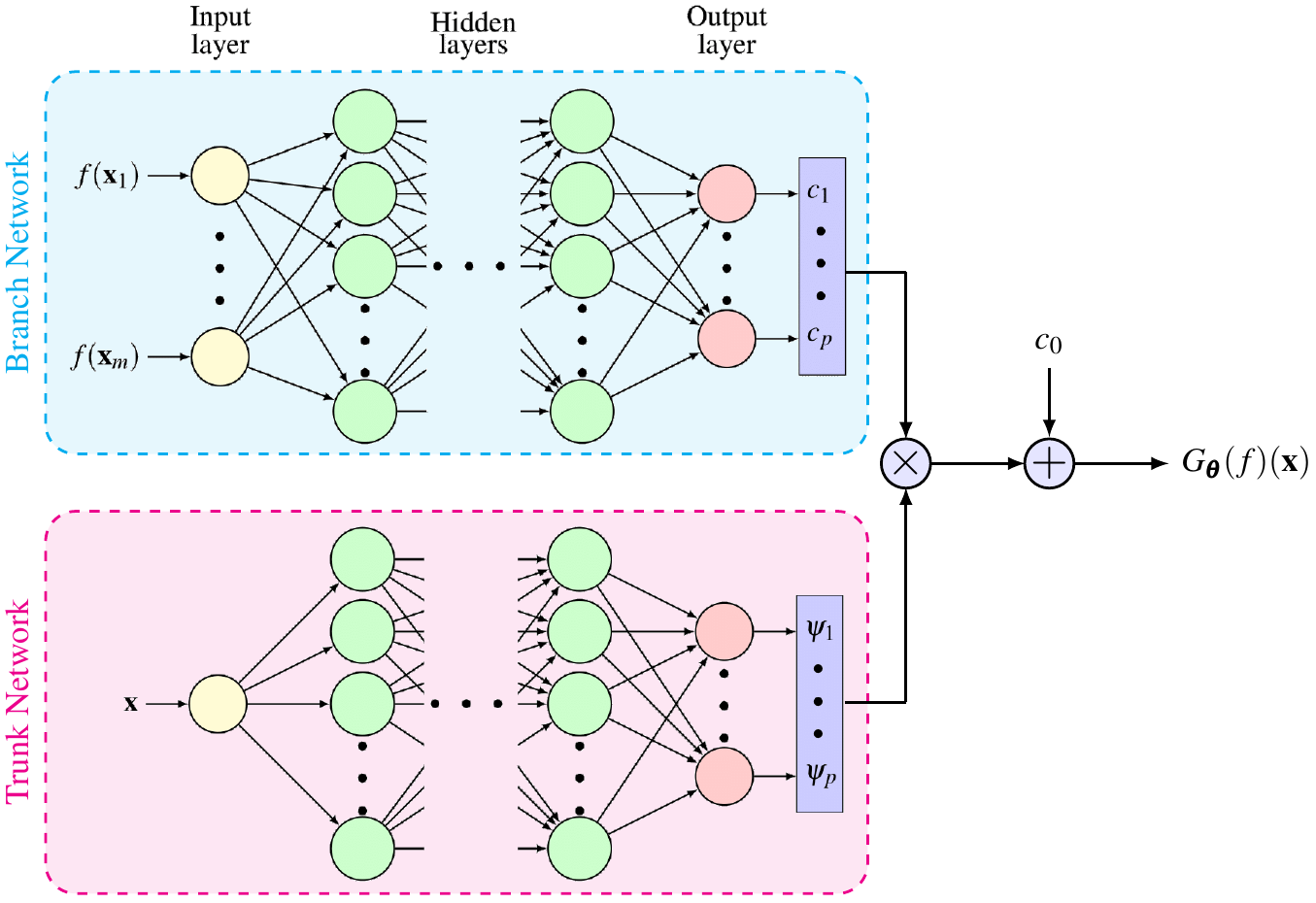}
	\caption{DeepONet architecture (stacked) \cite{lu2019deeponet} used in the present study. }\label{fig:deeponet} 
\end{figure} 
The DeepONet architecture proposed in Lu et al. \cite{lu2019deeponet} approximates the solution operator $G(f)(\xm)$ as 
\begin{equation} \label{eq:deep} 
    G_{\thetaa}(f)(\xm) = c_0 + \sum_{i=1}^p c_i(f(\xm_1),\dots,f(\xm_m);\thetaa_\mathrm{branch})\psi_i(\xm;\thetaa_\mathrm{trunk}), 
\end{equation}
where $c_0$ is a constant bias parameter; $c_i(\cdot),~i=1,\dots,p,$ are obtained from a neural network, known as the \textit{branch network}, that has the input function (\textit{e.g.}, external source term) $f(\cdot)$ measured at $\{\xm_i\}_{i=1}^m$ locations as input; $\psi_i(\cdot),~i=1,\dots,p,$ are obtained from a neural network, known as the \textit{trunk network}, that has the location $\xm$ as the input; and $\thetaa=[\thetaa_\mathrm{branch}^T \quad \thetaa_\mathrm{trunk}^T]^T$ is the parameter vector containing weights and biases $\thetaa_\mathrm{branch}$ and $\thetaa_\mathrm{trunk}$ from the \textit{branch} and \textit{trunk} networks, respectively, and $c_0$. This approximation is based on the universal approximation theorem in Chen and Chen \cite{chen1995universal} for operators. 
A schematic of the DeepONet architecture used in this study is shown in Figure \ref{fig:deeponet}. 
The main advantage of using a DeepONet architecture over other standard ones, such as feedforward, recurrent, etc., is that DeepONet provides a mesh-independent approximation of the solution operator since DeepONet uses a set of parameters $\thetaa$ trained only once with different mesh discretizations \cite{li2020fourier,lu2019deeponet}. 

\begin{figure}
    \centering
    \includegraphics[scale=1]{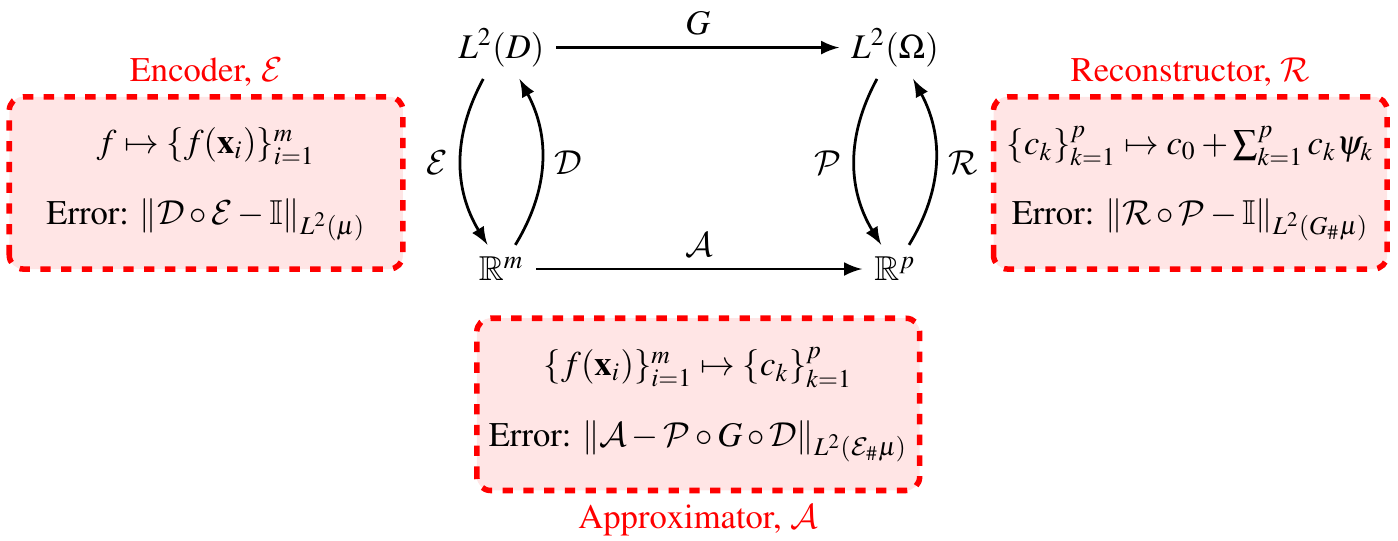}
    \caption{DeepONet components and corresponding errors adapted from Lanthaler et al. \cite{lanthaler2021error}.}
    \label{fig:deeponet_error}
\end{figure}

{
Lanthaler et al. \cite{lanthaler2021error} showed that the DeepONet can be thought of as a composition of three components. The first component is an \textit{encoder} $\mathcal{E}$ that maps the input function space to $m$ sensor inputs $\{f(\xm_i)\}_{i=1}^{m}$ for $\{\xm_i\}_{i=1}^{m}\subset D$. The second component, the \textit{approximator} $\mathcal{A}$, models the relation between $N_f$ sensor inputs $\{f(\xm_i)\}_{i=1}^{m}$ and the coefficients $\{c_k\}_{k=1}^p$. 
The encoder and the approximator together comprise the \textit{branch network}. The third or the final component of DeepONet is the \textit{reconstructor}, which maps the output of the branch network to the response $u(\xm)$, thereby showing dependence on the trunk network outputs $\{\psi_k\}_{k=1}^p$.
With approximate inverses of the encoder and reconstructor defined as \textit{decoder} $\mathcal{D}$ and projector $\mathcal{P}$, respectively, three errors associated with DeepONet approximation can be defined as \textit{encoding error} $\mathcal{E}_{e}=\lVert \mathcal{D}\circ\mathcal{E}-\mathbb{I} \rVert_{L^2(\mu)}$, \textit{approximation error} $\mathcal{E}_{a}=\lVert \mathcal{A} - \mathcal{P}\circ G\circ\mathcal{D} \rVert_{L^2(\mathcal{E}_{\#}\mu)}$, and \textit{reconstruction error} $\mathcal{E}_{r}=\lVert \mathcal{R}\circ\mathcal{P} -\mathbb{I} \rVert_{L^2(G_{\#}\mu)}$, where $\mathbb{I}$ is an identity matrix; $\mu$ is the measure associated with $f$; $\mathcal{E}_{\#}\mu$ and $G_{\#}\mu$ are push-forward measures associated with the \textit{encoder} and operator $G$, respectively. 
Figure \ref{fig:deeponet_error} shows the three components of a DeepONet and associated errors as defined here. 
The total approximation error associated with a DeepONet depends on these three errors as outlined in Theorem 3.2 in Lanthaler et al.~\cite{lanthaler2021error}. 
}

The parameter vector $\thetaa$ is estimated by solving the optimization problem 
\begin{equation} \label{eq:opt}
\begin{split}
 \min\limits_{\thetaa} J(\thetaa) :&= \frac{1}{N_fN_d} \sum_{j=1}^{N_f}\sum_{i=1}^{N_d} \lvert G_{\thetaa}(f_j)(\xm_i) - G(f_j)(\xm_i) \rvert^2\\
 &= \frac{1}{N_fN_d} \sum_{j=1}^{N_f}\sum_{i=1}^{N_d} \left\lvert c_0 + \sum_{k=1}^p c_k(f_j(\xm_1),\dots,f_j(\xm_m);\thetaa_\mathrm{branch})\psi_k(\xm_i) - G(f_j)(\xm_i;\thetaa_\mathrm{trunk}) \right\rvert^2,\\
\end{split} 
\end{equation} 
where in the training dataset for each of $N_f$ realizations of the external source term $\{f_j(\xm)\}_{j=1}^{N_f}$, $N_d$ measurements of the response at locations $\{\xm_i\}_{i=1}^{N_d}$, which might not be the same as the locations where the external source term is measured, are used. These sensor locations can be random or equispaced inside the domain and kept fixed during training. The optimization problem in \eqref{eq:opt} can be solved using stochastic gradient descent \cite{bottou2010large,bottou2012stochastic,de2020topology} to obtain optimal parameter values of the DeepONets. In this paper, we use a popular variant of stochastic gradient descent, known as the Adam algorithm \cite{kingma2014adam}, to train DeepONets. Note that in Lu et al. \cite{lu2019deeponet}, the input function $f(\xm)$ is sampled from function spaces with the locations $\xm$ as domain and the external source $f$ as range, such as Gaussian random field and Chebyshev polynomials. However, for practical reasons, in this paper, we use probability distribution of the uncertain variables associated with the problems to sample the input function. 
However, the number of training data for DeepONets can be limited if the computational cost of the simulation of the physical system or the cost of repeating the experiments is significant. In this study, to address the challenge of training DeepONets in the presence of limited data, we propose a bi-fidelity training approach utilizing a computationally inexpensive \textit{low-fidelity} model of the physical system. 
In the next section, we detail this proposed approach and discuss two applications to address parametric and structural uncertainties present in the system.

\section{Methodology} \label{sec:method}

In the present work, we assume a low-fidelity model of the physical system in \eqref{eq:gov} is given by
\begin{equation} \label{eq:u_l}
    \mathcal{L}_l(u_l)(\xm) = f_l(\xm), \qquad \xm \in \Omega,
\end{equation}
where $u_l(\xm)$ is the low-fidelity response; $\mathcal{L}_l$ is the differential operator related to the low-fidelity model; and $f_l(\xm)$ is the external source in the low-fidelity model. 
Next, we denote the discrepancy between the solution of the low-fidelity model $u_l(\xm)$ and that of the physical system $u(\xm)$ as 
\begin{equation} \label{eq:u_c}
    u_d(\xm) = u(\xm) - u_l(\xm).
\end{equation} 
As it is often the case that the low-fidelity model will capture many important aspects of the physical system's response, we assume that the training of a DeepONet for modeling $u_d(\xm)$ is easier than modeling the true response $u$ and that generating the low-fidelity response $u_l(\xm)$ is computationally inexpensive. 
We present two applications of the proposed approach next. 
Note that 
$u_d(\xm)$ is estimated from the DeepONet and $u_l(\xm)$ from the low-fidelity model for any new $\xm\in\Omega$ to predict the response $u(\xm)$. Further, this new location $\xm$ need not be the same locations used during training or the places where the external source is measured due to the use of DeepONet to model $u_d(\xm)$. 
We also do not use another DeepONet or neural network to model the low-fidelity response as the simulation of the low-fidelity model is computationally inexpensive, and this avoids any approximation error introduced through a low-fidelity trained network.

  \begin{figure}[!htb]
        \centering
\includegraphics[scale=1]{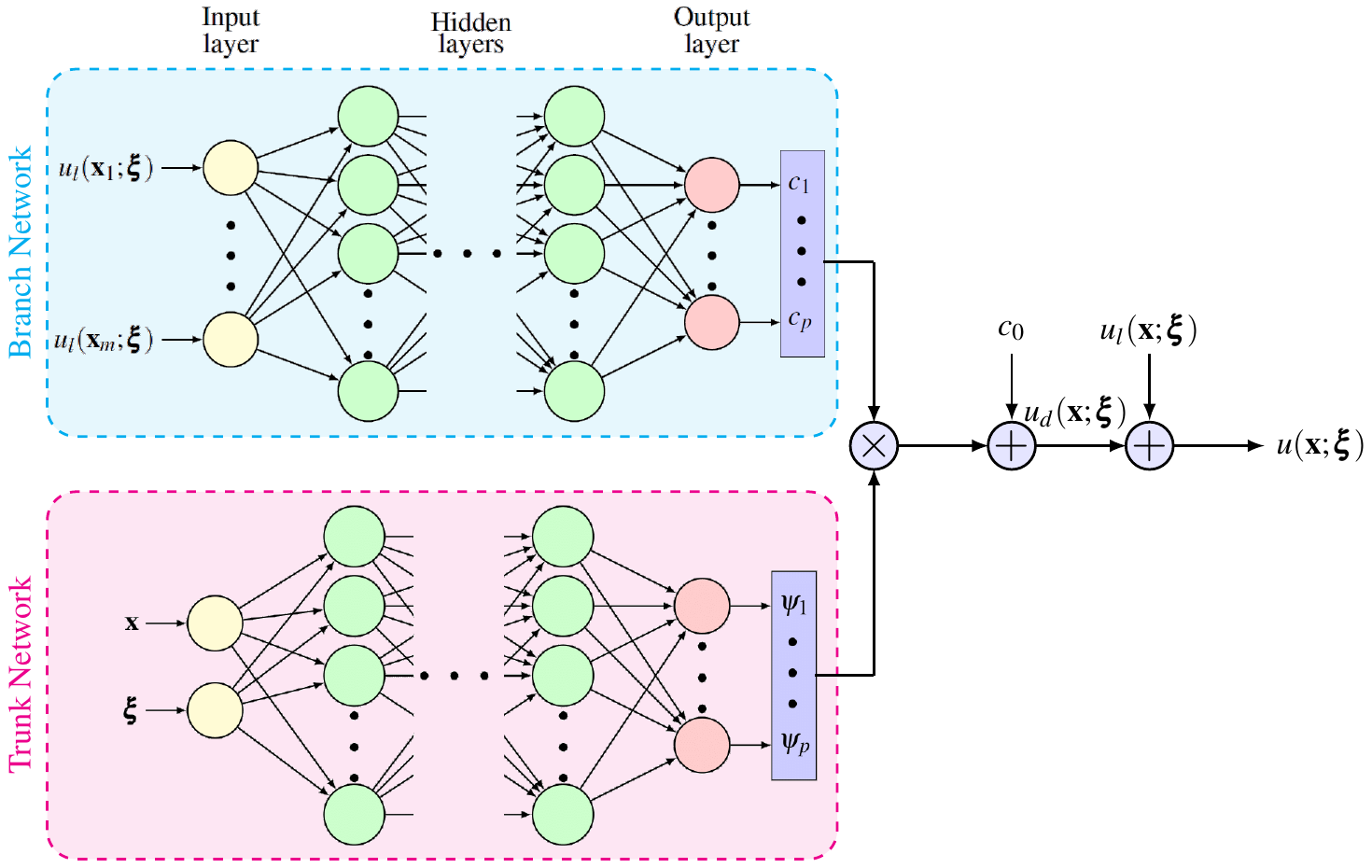}
    \caption{DeepONet used for bi-fidelity modeling of uncertain systems in Section \ref{sec:app1}. }
    \label{fig:Uncertain_systems}
\end{figure}

\subsection{Application I: Systems with Parametric Uncertainty} \label{sec:app1} 
The proposed approach can be used for modeling systems with parametric uncertainty, where a computationally inexpensive low-fidelity model of a similar system is available. 
For example, consider a system described by  
\begin{equation} \label{eq:app1}
    \left(\mathcal{L}_d  + \mathcal{L}_s(\xii)\right) (u)(\xm;\xii) = f(\xm;\xii), 
\end{equation}
where $\xii$ is the vector of uncertain variables; $\mathcal{L}_d$ and $\mathcal{L}_s$ are deterministic and uncertain (or stochastic) parts of the differential operator in \eqref{eq:gov}, respectively; the response $u$ depends on the uncertain variables $\xii$; and the external source may depend on $\xii$ as well. 
In this case, a low-fidelity model for \eqref{eq:app1} can be defined as follows
\begin{equation} \label{eq:u_l_app1}
    \mathcal{L}_d (u_l)(\xm;\xii) = f_l(\xm). 
\end{equation} 
Hence, the response of the physical system is approximated as 
\begin{equation} \label{eq:deeponet_app1} 
\begin{split}
    u(\xm;\xii) &= u_l(\xm;\xii) + u_d(\xm;\xii) \\
    &\approx u_l(\xm;\xii) + G_{\thetaa}(u_l)(\xm,\xii) \\
    &= u_l(\xm;\xii) + c_0 + \sum_{i=1}^p c_i(u_l(\xm_1;\xii),\dots,u_l(\xm_m;\xii);\thetaa_\mathrm{branch})\psi_i(\xm,\xii;\thetaa_\mathrm{trunk}), 
\end{split}
\end{equation}
where $\left\{u_l(\xm_j;\xii)\right\}_{j=1}^m$ are used as the input of the branch network and $(\xm,\xii)$ as the input of the trunk network to model the discrepancy $u_d(\xm;\xii)$. A schematic of the approximation in \eqref{eq:deeponet_app1} is shown in Figure \ref{fig:Uncertain_systems}. 

As a special case, when the operators $\mathcal{L}_d$ and $\mathcal{L}_s(\xii)$ are linear, using \eqref{eq:u_c} in \eqref{eq:app1} we arrive at 
\begin{equation}
    \begin{split}
        u_d(\xm;\xii) = -\left( \mathcal{L}_d + \mathcal{L}_s(\xii) \right)^{-1}\left( \mathcal{L}_s(\xii)(u_l)(\xm)-f_u(\xm;\xii) \right),
    \end{split} 
\end{equation} 
which is modeled using a DeepONet, where $u_l(\xm)$ is the solution to \eqref{eq:u_l_app1} and $f_u(\xm;\xii)=f(\xm;\xii)-f_l(\xm)$. Note that we do not necessarily require the linearity of the differential operators for the proposed approach to apply. However, we make that assumption to illustrate the simplified construction of the low-fidelity model and the correction. 

\begin{figure}[!htb]
    \centering
\includegraphics[scale=1]{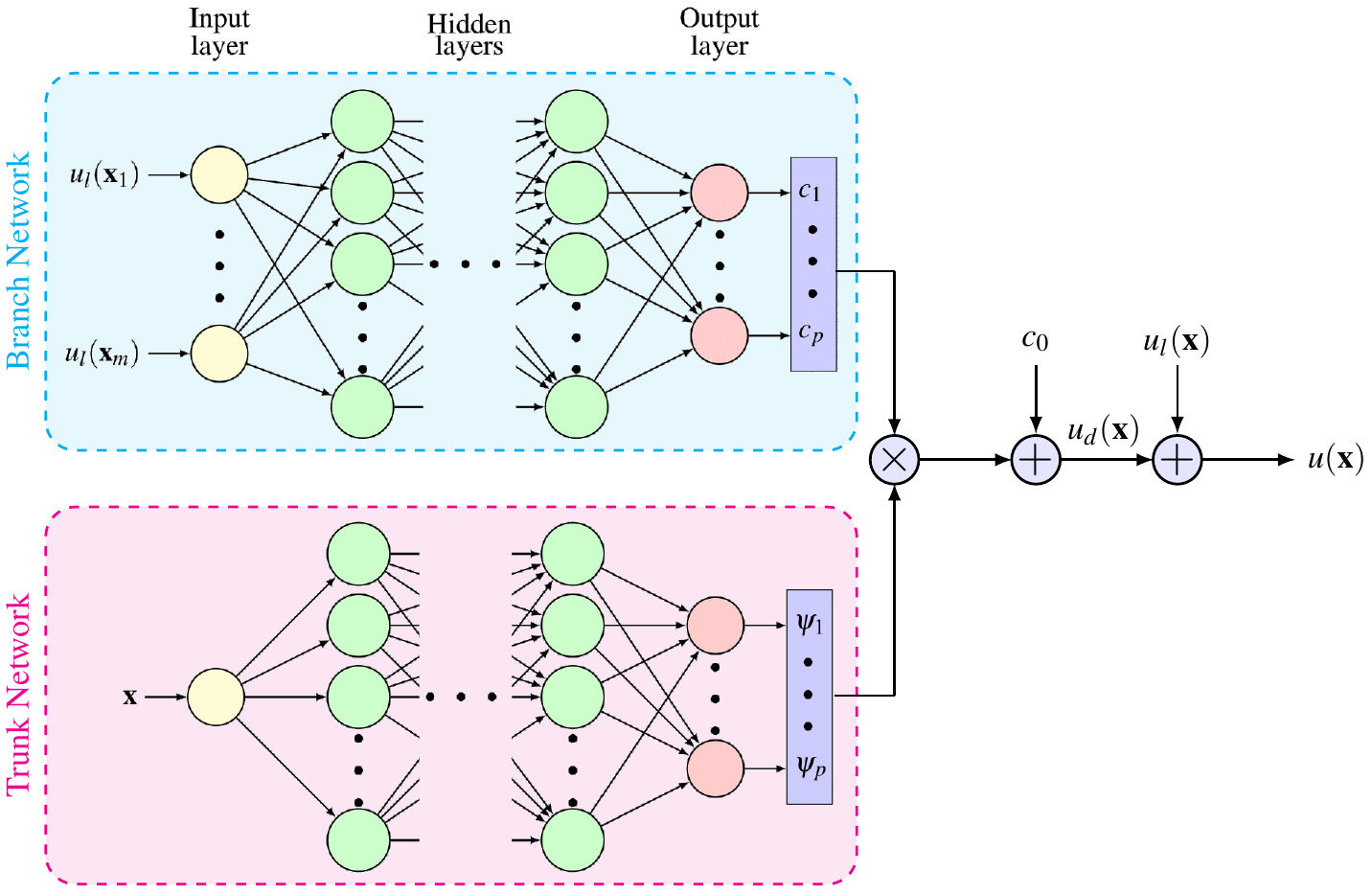}
           \caption{DeepONet used for bi-fidelity modeling of partially unknown systems in Section \ref{sec:app2}. }
        \label{fig:Unknown_systems}
    \end{figure}

\subsection{Application II: Systems with Structural Uncertainty} \label{sec:app2}
In this application, we propose to use \eqref{eq:u_c} to model the response of a partially unknown system. For example, let us assume we can rewrite \eqref{eq:gov} as follows 
\begin{equation} \label{eq:app2}
    \left(\mathcal{L}_k + \mathcal{L}_u\right)(u)(\xm) = f_k(\xm) + f_u(\xm), 
\end{equation}
where $\mathcal{L}_k$ and $\mathcal{L}_u$ are the known and unknown parts of the  differential operator, respectively; and $f_k$ and $f_u$ are the known and unknown parts of the external source, respectively. A low-fidelity model for this application can be defined as 
\begin{equation} \label{eq:u_l_app2}
    \mathcal{L}_k(u_l)(\xm) = f_k(\xm). 
\end{equation} 
The response of the physical system is approximated next as 
\begin{equation} \label{eq:deeponet_app2}
\begin{split}
    u(\xm) &= u_l(\xm) + u_d(\xm) \\
    &\approx u_l(\xm) + G_{\thetaa}(u_l)(\xm) \\
    &= u_l(\xm) + c_0 + \sum_{i=1}^p c_i(u_l(\xm_1),\dots,u_l(\xm_m);\thetaa_\mathrm{branch})\psi_i(\xm;\thetaa_\mathrm{trunk}),
\end{split}
\end{equation}
where we use $\left\{u_l(\xm_j)\right\}_{j=1}^m$ as the branch network input and $\xm$ as the trunk network input. Figure \ref{fig:Unknown_systems} shows a schematic of the approximation in \eqref{eq:deeponet_app2}. 

As a special case, assuming the differential operators $\mathcal{L}_k$ and $\mathcal{L}_u$ are linear leads to 
the following relation for $u_d(\xm)$ in \eqref{eq:u_c} 
\begin{equation}
    \begin{split}
        u_d(\xm) = -\left( \mathcal{L}_k + \mathcal{L}_u \right)^{-1}\left( \mathcal{L}_u(u_l)(\xm)-f_u(\xm) \right). 
    \end{split}
\end{equation}
As before, we do not necessarily require linear differential operators to apply the proposed method. Note that it is straightforward to apply the proposed approach for an uncertain and partially unknown system using $\left(\xm,\xii\right)$ as trunk network input, as in the previous subsection and addresses both parametric and structural uncertainties. Also, the low-fidelity model can be uncertain in these two applications, but for brevity we do not explicitly write the dependence of the low-fidelity solution on $\xii$ in this section. 

\subsection{Choice of Low-fidelity Model} 
{
The choice of the low-fidelity model can have a significant effect on the total DeepONet approximation error. 
As discussed in Section \ref{sec:back}, with a three-component decomposition of the DeepONet, Lanthaler et al. \cite{lanthaler2021error} showed that the total error in a DeepONet approximation depends on errors in each of these components \cite[Eq.~(3.4)]{lanthaler2021error}. Among them, the error in the \textit{approximator} depends on the neural network approximation error, such as in Yarotsky \cite[Thm.~1]{yarotsky2017error}.  
On the other hand, the errors in \textit{encoder} and \textit{reconstructor} depend on the eigenvalues of covariance operator of the branch network inputs and $G(f)(\xm,\xii)$ \cite{lanthaler2021error}, respectively. In particular, these errors can be bounded from below using $\sqrt{\sum_{i>p}\lambda_i}$, where $\lambda_1\geq\lambda_2\geq\dots$ are eigenvalues of the covariance operator of the branch network input for \textit{encoding error} $\mathcal{E}_{e}$ and eigenvalues of the covariance operator of $G(f)(\xm,\xii)$ for \textit{reconstruction error} $\mathcal{E}_{r}$, respectively. 

In the bi-fidelity DeepONet, we are replacing the external source term with the low-fidelity response in the branch network input, and the discrepancy $u_d(\xm;\xii)$ is modeled using the DeepONet. 
Hence, an ideal choice for the low-fidelity model that will reduce the \textit{encoding error} $\mathcal{E}_{e}$ will be a low-pass filter system that will remove most of the eigenvalues for $i>p$ of the covariance operator of the branch network input. Also, in many cases, encoding the low-fidelity model's response will be easier than encoding the external source reducing the \textit{encoding error}. 
Similarly, to reduce the \textit{reconstruction error} $\mathcal{E}_{r}$, we can select a low-fidelity model such that the eigenvalues of covariance of $u_d(\xm,\xii)$ has smaller magnitudes for $i>p$. 
However, in practice, we may be limited in terms of choosing the low-fidelity model. A simplified physics or a coarser discretization is often used as the low-fidelity model in most of the multi-fidelity methods. In our numerical examples described in Section \ref{sec:ex}, we will show that these practical low-fidelity models can still provide significant improvements in the total DeepONet approximation error.

As an elementary example to illustrate the above discussion, we consider a duffing oscillator with governing differential equation 
\begin{equation} \label{eq:duff} 
    \frac{\mathrm{d}^2u(t;\xii)}{\mathrm{d}t^2} + \delta \frac{\mathrm{d}u(t;\xii)}{\mathrm{d}t} + \alpha u(t;\xii) + \beta u^3(t;\xii) = \gamma \cos{\omega t}; \quad t\in [0,10],~u(0) = 0, 
\end{equation} 
where $\delta$ is the damping coefficient; $\alpha$ is the linear stiffness coefficient; $\beta$ is the nonlinear stiffness coefficient;  $\gamma$ is the amplitude of the external force; $\omega$ is the frequency of the external force; and $u_0$ is the initial response of the system. We consider the two applications of bi-fidelity DeepONet discussed in Section \ref{sec:method}. In the first application, the parameters $\beta$ and $\omega$ are assumed as uncertain, and a low-fidelity model that does not include the term $\beta u^3(t;\xii)$ is considered. For this case, Figure \ref{fig:ExI_LB1} shows the lower bound of the \textit{reconstruction error} $\mathcal{E}_{r}$ computed using $\sqrt{\sum_{i>p}\lambda_i}$, where $\lambda_1\geq\lambda_2\geq\dots$ are eigenvalues of the sample covariance matrix of $u_d(\xm,\xii)$ estimated using 2000 realizations of the random parameters  \cite[Th.~3.4]{lanthaler2021error}. In the second application, we use a low-fidelity model that does not consider the term $\beta u^3(t;\xii)$ and $\delta$ is replaced with a different $\delta_\mathrm{nom}$. The lower bound of the \textit{reconstruction error} $\mathcal{E}_{r}$ computed similarly for this case is shown in Figure \ref{fig:ExI_LB2}. 
These two plots show that for smaller $p$ in \eqref{eq:deep}, the bi-fidelity DeepONet provides significantly lower bound for the \textit{reconstruction error}. For large $p$, this lower bound for both bi-fidelity DeepONet and the DeepONet described in Section \ref{sec:back}, denoted as \textit{standard} in the rest of the paper, reaches zero. 

\begin{figure}[!htb] 
    \centering
    \centering
    \begin{subfigure}[t]{0.5\textwidth}
        \centering
\includegraphics[scale=1]{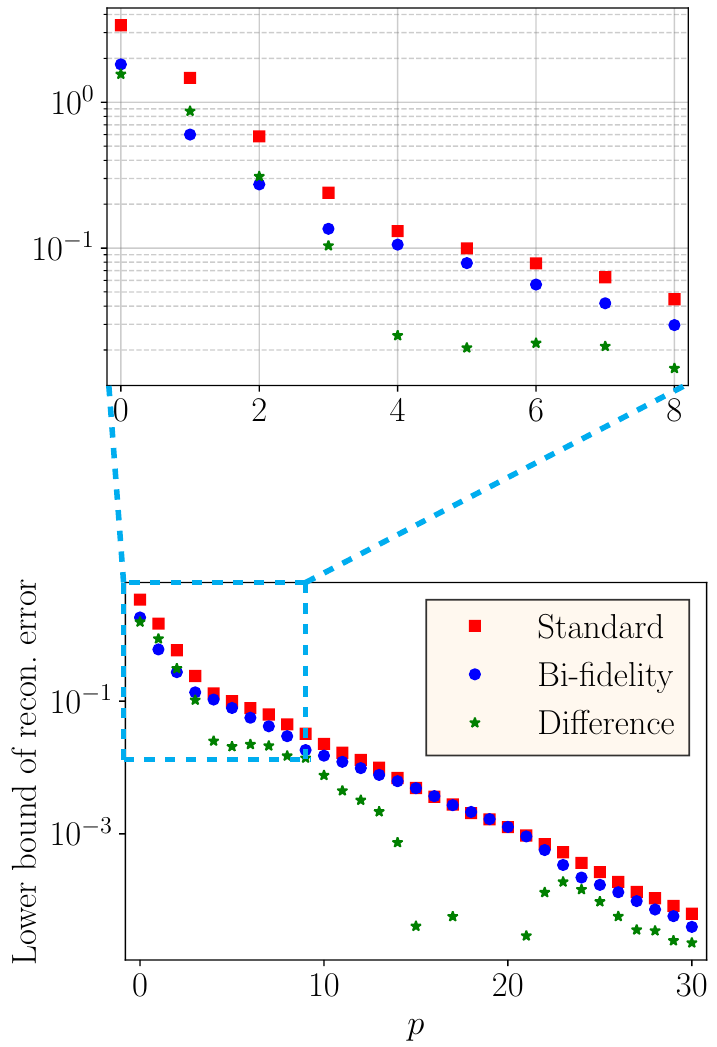}
    \caption{Application I: Parametric uncertainty}
    \label{fig:ExI_LB1} 
    \end{subfigure}%
    ~ 
    \begin{subfigure}[t]{0.5\textwidth}
        \centering 
\includegraphics[scale=1]{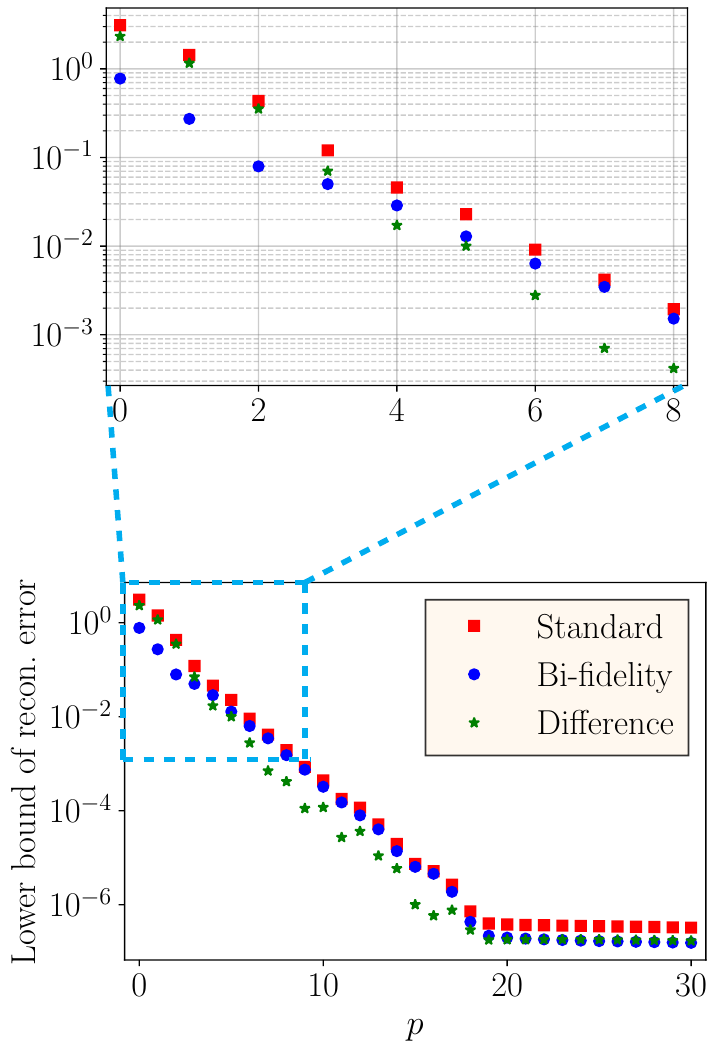}
    \caption{Application II: Structural uncertainty}
    \label{fig:ExI_LB2}
    \end{subfigure} 
    \caption{The lower bounds of \textit{reconstruction error} using standard DeepONet, Bi-fidelity DeepONet, and their difference as estimated using Lanthaler et al. \cite[Th.~3.4]{lanthaler2021error} for a duffing oscillator used in Example \ref{sec:ex1}. }
    \label{fig:ExI_LB}
\end{figure}

Experimental results for these two applications of bi-fidelity DeepONet are provided in the next section. Note that, we can also choose different low-fidelity models that do not necessarily provide a smaller lower bound for the \textit{reconstruction error}, but the \textit{approximation} and \textit{encoding errors} are smaller with bi-fidelity DeepONet compared to a standard DeepONet as is done in Example \ref{sec:ex2}. 
}
\section{Numerical Examples} \label{sec:ex} 
In this section, we illustrate the proposed approach in approximating the discrepancy between the low-fidelity response and the physical system's response with DeepONets using three numerical examples. The first example uses nonlinear ordinary differential equation for an initial value problem. In the second example, we model a nonlinear heat transfer in a thin plate, where a partial differential equation for the boundary problem is used. We consider both parametric and structural uncertainties to illustrate the applications I and II of the proposed approach from Section \ref{sec:method} in these two examples. The third example considers flow through a wind farm with a combination of parametric and structural uncertainties, where two-dimensional and three-dimensional simulations are used as the \textit{low-} and \textit{high-fidelity} model of the  system, respectively. In these examples, we use two separate datasets, namely a training dataset $\Dtr=\left\{(\xm_i,\xii_i),u(\xm_i;\xii_i),\{f^{(i)}(\xm_j)\}_{j=1}^m\right\}_{i=1}^{\Ntr}$ and a validation dataset $\Dval=\left\{(\xm_i,\xii_i),u(\xm_i;\xii_i),\{f^{(i)}(\xm_j)\}_{j=1}^m\right\}_{i=1}^{\Nval}$. 
We use the training dataset $\Dtr$ to train the DeepONets, and the accuracy of the trained DeepONets is measured using the validation dataset $\Dval$, with the relative validation root mean squared error (RMSE), $\varepsilon_\mathrm{val}$, defined as follows 
\begin{equation} \label{eq:val_rmse} 
    \epsv = \frac{\lVert \um_\mathrm{pred} - \um_\mathrm{val} \rVert_2}{\lVert\um_\mathrm{val}\rVert_2}, 
\end{equation} 
where $\lVert\cdot\rVert_2$ is the $\ell_2$-norm of its argument, $\um_\mathrm{pred}$ is the vector of predicted response using DeepONets, and $\um_\mathrm{val}$ is from the validation dataset $\Dval$. 
Note that we select the number of hidden layers in the trunk and branch networks and the number of neurons per hidden layer using an iterative procedure, where we increase the number of neurons per hidden layer gradually up to an upper limit (100 in this study) and then add hidden layers to the network until we observe no reduction of $\epsv$. For the number of terms $P$ in \eqref{eq:deeponet_app1} and \eqref{eq:deeponet_app2}, we increase $P$ in multiples of 10 to a maximum, which is the number of neurons in the hidden layers, to observe any reduction of $\epsv$. 

\subsection{Example 1: Duffing Oscillator} \label{sec:ex1} 

In this example, we use a duffing oscillator described by the governing differential equation \eqref{eq:duff}. We use a neural network that has three hidden layers with 40 neurons each as the branch network and a neural network that has two hidden layers with 40 neurons each as the trunk network. We use ELU as the activation function in these networks \cite{clevert2015fast}. The number of terms in the summation in \eqref{eq:deep} is fixed at $p=20$ following the procedure discussed in the previous section. 

\subsubsection{Application I: Parametric Uncertainty} \label{sec:duff_app1} 
\begin{figure}[!htb] 
    \centering
    \centering
    \begin{subfigure}[t]{0.5\textwidth}
        \centering
        \includegraphics[scale=1]{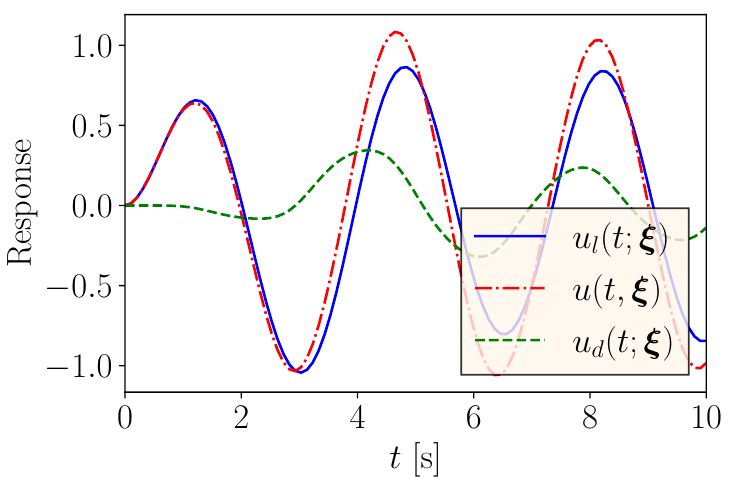}
    \caption{Comparison of the low-fidelity response $u_l(t;\xii)$, correction $u_d(t;\xii)$, and true response $u(t;\xii)$ for one realization of the uncertain variables }
    \label{fig:ExI_u_c1} 

    \end{subfigure}%
    ~~~~ 
    \begin{subfigure}[t]{0.5\textwidth}
        \centering 
\includegraphics[scale=1]{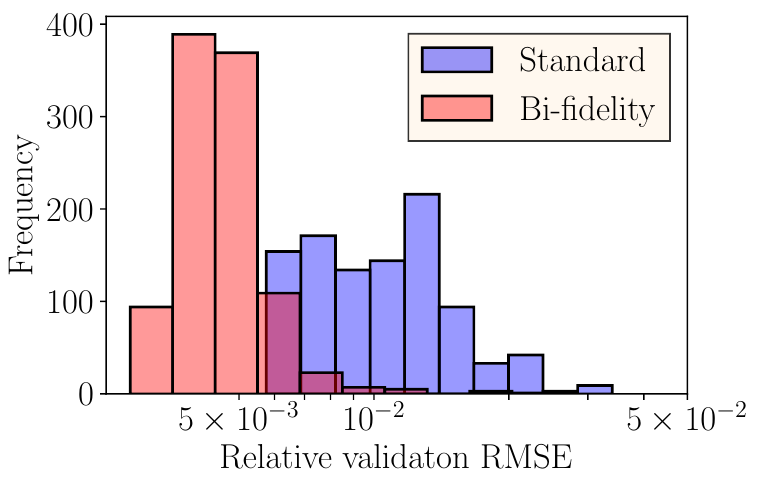}
    \caption{Relative validation error using DeepONets }
    \label{fig:ExI_hist1}
    \end{subfigure} 
    \caption{The use of DeepONets to model the response of an uncertain Duffing oscillator. Here, a standard use of DeepONet as described in Section \ref{sec:back} is denoted as `Standard', and the proposed use of DeepONet to model the discrepancy between the low-fidelity model and true system is denoted as `Bi-fidelity.'} 
    \label{fig:ExI_App1}
\end{figure}

For the oscillator defined in \eqref{eq:duff}, we use $\alpha=1.0$, $\delta=1.0$, and $\gamma=2.5$. 
The uncertain variables are assumed as $\xii=(\beta,\omega)$ with $\beta\sim\mathcal{U}[0.5,2.5]$ and $\omega\sim\mathcal{U}[1.5,2.5]$, where $\mathcal{U}[a,b]$ denotes a uniform distribution between limits $a$ and $b$. We use a training dataset $\Dtr$ with $\Ntr=200$ and a validation dataset $\Dval$ with $\Nval=1000$ realizations of the uncertain variables uniformly sampled from their corresponding distributions, respectively, where the response is measured at 100 equidistant points along $t$. 
A low-fidelity model of the system can be specified without the nonlinear stiffness, as follows 
\begin{equation} 
     \frac{\mathrm{d}^2u_l(t;\xii)}{\mathrm{d}t^2} + \delta \frac{\mathrm{d}u_l(t;\xii)}{\mathrm{d}t} + \alpha u_l(t;\xii) = \gamma \cos{\omega t}; \quad t\in [0,10],~u_l(0) = 0,
\end{equation} 
where $u_l(t;\xii)$ is the low-fidelity response. Note that uncertainty in the low-fidelity model only comes through the uncertain variable $\omega$. 
Figure \ref{fig:ExI_u_c1} depicts a representative realization of  $u(t;\xii)$, the low-fidelity response $u_l(t;\xii)$, and the discrepancy $u_d(t;\xii)$. 
Note that in this example the cost of generating the low- and high-fidelity datasets is insignificant. 

The proposed use of DeepONet to model the discrepancy $u_d(t;\xii)$ between the low-fidelity and the true responses produces $4.72\times10^{-3}$ as the validation error $\varepsilon_\mathrm{val}$ with the validation dataset $\Dval$. In contrast, the standard DeepONet results in a validation error of $1.11\times10^{-2}$. Figure \ref{fig:ExI_hist1} compares the histograms of the relative validation RMSE of these two approaches. Hence, in this case, the proposed approach improves the validation error by more than a factor of two, as modeling of the discrepancy $u_d(t;\xii)$ proved to be easier using the same DeepONet architecture than modeling the response $u(t;\xii)$.

\begin{figure}[!htb] 
    \centering
    \centering
    \begin{subfigure}[t]{0.5\textwidth}
        \centering
        \includegraphics[scale=1]{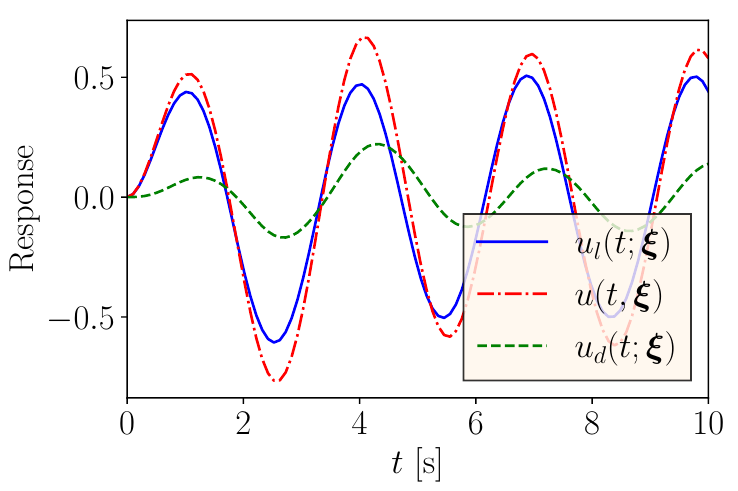}
    \caption{Comparison of the low-fidelity response $u_l(t;\xii)$, correction $u_d(t;\xi)$, and true response $u(t;\xi)$ for one realization of the uncertain variables }
    \label{fig:ExI_u_c2} 

    \end{subfigure}%
    ~ 
    \begin{subfigure}[t]{0.5\textwidth}
        \centering 
\includegraphics[scale=1]{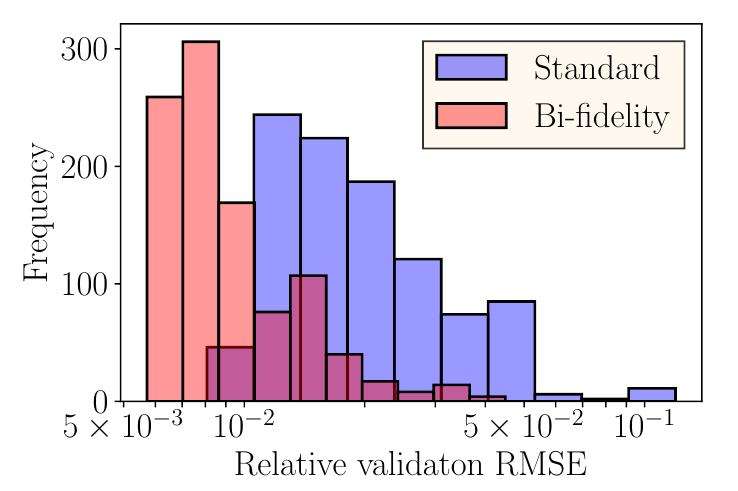}
    \caption{Relative validation error using DeepONets }
    \label{fig:ExI_hist2}
    \end{subfigure} 
    \caption{The use of DeepONets to model the response of an partially unknown Duffing oscillator in Example 1.} 
    \label{fig:ExI_App2}
\end{figure} 

\subsubsection{Application II: Partially Unknown System (Structural Uncertainty)} 
In this case, we assume $\alpha=1.0$, $\delta=1.0$, and $\gamma=2.5$ as before, but $\beta=1.0$. The frequency of the external load is assumed to be uniformly distributed as before. We use a training dataset $\Dtr$ with $\Ntr=200$, where we also add 5\% additive Gaussian white noise to the dataset, and a validation dataset $\Dval$ with $\Nval=1000$. The low-fidelity model is assumed as
\begin{equation}
     \frac{\mathrm{d}^2u_l(t;\xii)}{\mathrm{d}t^2} + \delta_\mathrm{nom} \frac{\mathrm{d}u_l(t;\xii)}{\mathrm{d}t} + \alpha u_l(t;\xii) = \gamma \cos{\omega t}; \quad t\in [0,10],~u_l(0) = 0,
\end{equation} 
where the nominal value of the damping coefficient is $\delta_\mathrm{nom}=1.5$. Hence, the unknown dynamics not modeled in the low-fidelity model can be given by 
\begin{equation}
    \mathcal{L}_u u(t;\xi) \equiv (\delta-\delta_\mathrm{nom}) \frac{\mathrm{d}u(t;\xi)}{\mathrm{d}t} + \beta u^3(t;\xi). 
\end{equation} 
Figure \ref{fig:ExI_u_c2} shows a comparison of the response from the low-fidelity model $u_l(t;\xii)$, true response $u(t;\xi)$, and their difference $u_d(t;\xi)$. 
When DeepONet is used to model the discrepancy $u_d(t,\xi)$ between the true response and the low-fidelity model response, it produces a validation error $\varepsilon_\mathrm{val}$ of $9.98\times10^{-3}$. Note that the standard use of DeepONet produces an $\varepsilon_\mathrm{val}$ of $2.22\times10^{-2}$ with the same training and validation datasets. Figure \ref{fig:ExI_hist2} shows a comparison of histograms from the proposed use of DeepONet to model $u_d(t;\xi)$ and the standard use of DeepONet as described in Section \ref{sec:back}. The figure shows that, in this case as well, modeling $u_d(t,\xi)$ as a function of the low-fidelity response $u_l(t;\xii)$ using DeepONet can achieve better accuracy compared to the standard use. 
Hence, this example shows the effectiveness of the proposed approach to model the discrepancy using DeepONet for uncertain and partially unknown dynamical systems, which reduces the validation error by a factor of more than two. 



\subsection{Example 2: Nonlinear Steady-state Heat Transfer in a Thin Plate} \label{sec:ex2}
\begin{figure}[!htb] 
    \centering
    \begin{subfigure}[t]{0.45\textwidth}
        \centering 
\includegraphics[scale=1]{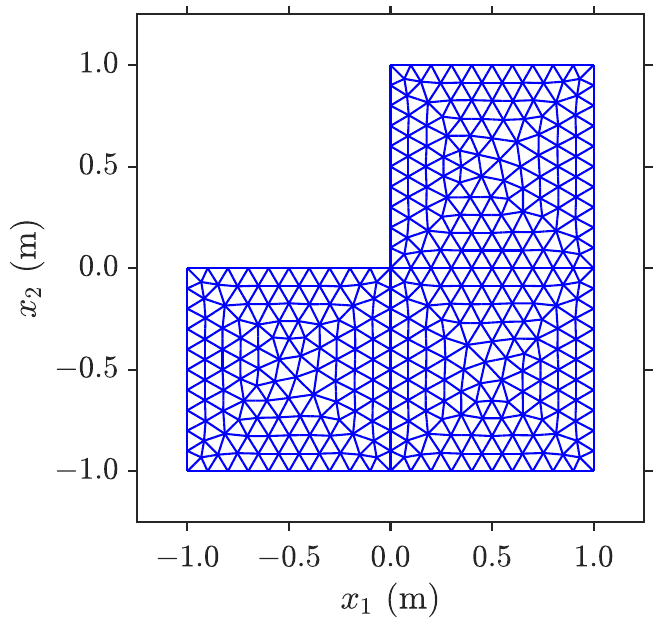}
    \caption{Finite element mesh used }
    \label{fig:ExII_mesh}
    \end{subfigure} ~
     \begin{subfigure}[t]{0.45\textwidth}
        \centering 
\includegraphics[scale=1]{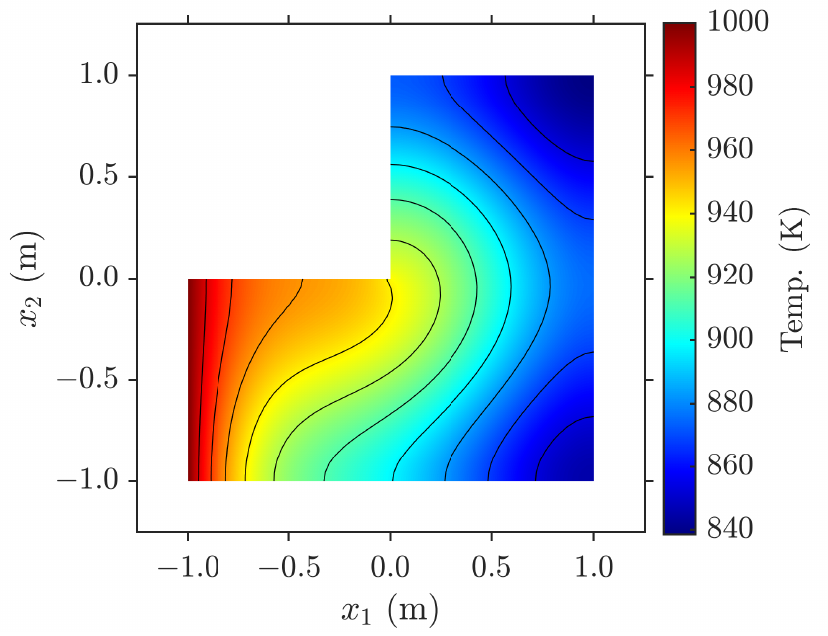}
    \caption{Temperature for one realization of the uncertain variables }
    \label{fig:ExII_temp_dist} 
    \end{subfigure} 
    \caption{Nonlinear steady-state heat transfer in a thin plate is studied in Example 2. } 
    \label{fig:ExII_App1}
\end{figure}

In our second example, we use a two-dimensional steady-state heat transfer problem in a thin L-shaped plate with dimensions shown in Figure \ref{fig:ExII_mesh}. The left boundary is assumed to have a constant temperature of $100$ K. The steady-state temperature at any location inside the plate, $T(\xm)$, is given by 
\begin{equation} \label{eq:heat} 
    -h_t\nabla(k(\xm)\nabla T(\xm)) + 2Q_c(\xm) + 2Q_r(\xm)= Q_i(\xm),
\end{equation}
where $\nabla:=\frac{\partial}{\partial x_1}+\frac{\partial}{\partial x_2}$; $h_t$ is the thickness of the plate; $k(\xm)$ is the thermal conductivity of the plate; $Q_c(\xm)$ is the amount of heat transferred through one face of the plate per unit area using convection; $Q_r(\xm)$ is the amount of heat transferred through one face of the plate per unit area using radiation; and $Q_i(\xm)$ is the amount of heat added to the plate per unit area from an internal heat source. 
Heat transferred by convection can be given by Newton's law of cooling as 
\begin{equation} \label{eq:newton_cooling}
    Q_c(\xm)=h_c(T(\xm)-T_a),
\end{equation}
where $h_c$ is the specified convection coefficient and $T_a$ is the ambient temperature. For heat transferred due to radiation, the Stefan-Boltzmann law can be used,
\begin{equation} \label{eq:stefan-Boltzmann}
    Q_r(\xm) = \epsilon \sigma (T^4(\xm)-T_a^4), 
\end{equation}
where $\epsilon$ is the emissivity of the plate surface and $\sigma$ is the Stefan-Boltzmann constant of the plate. Given \eqref{eq:newton_cooling} and \eqref{eq:stefan-Boltzmann}, we rewrite \eqref{eq:heat} as 
\begin{equation} \label{eq:ExII_gov_eqn}
    -h_t\nabla(k(\xm)\nabla T(\xm)) + 2h_cT(\xm) + 2\epsilon\sigma T^4(\xm) = 2h_cT_a + 2\epsilon\sigma T_a^4 + Q_i(\xm). 
\end{equation} 
In this example, we set $h_t=0.01$ m; $h_c=1$ $\text{W(m}^2\text{K)}^{-1}$; $\epsilon=0.025$; and $\sigma=5.670373\times10^{-8}$ $\text{W(m}^2\text{K}^4\text{)}^{-1}$ (similar to copper) to generate the datasets. We also assume the internal heat source to be an uncertain squared exponential source centered at the $(0,0)$ corner given by 
\begin{equation} \label{eq:Q}
    Q_i(\xm) = Q_0 (1+0.1\xi_Q)\exp\left(-\sum_{i=1}^2x_i^2\right), 
\end{equation}
where $Q_0=5\times10^3$ $\text{Wm}^{-2}$ and $\xi_Q$ is a standard Gaussian random variable. 
We use the finite element method with the mesh shown in Figure \ref{fig:ExII_mesh} to solve \eqref{eq:ExII_gov_eqn}. Figure \ref{fig:ExII_temp_dist} shows a realization of the steady-state temperature in the plate. 
In this example, we use a neural network that has three hidden layers with 40 neurons each and ELU activation as the branch network, and a neural network that has two hidden layers with 40 neurons each and ELU activation as the trunk network. The number of terms in the summation in \eqref{eq:deep} is fixed at $p=20$ to produced the smallest validation error. 

\subsubsection{Application I: Uncertain Thermal Conductivity} 
\begin{figure}[!htb]
    \centering
    \begin{subfigure}[t]{0.45\textwidth}
        \centering
        \includegraphics[scale=1]{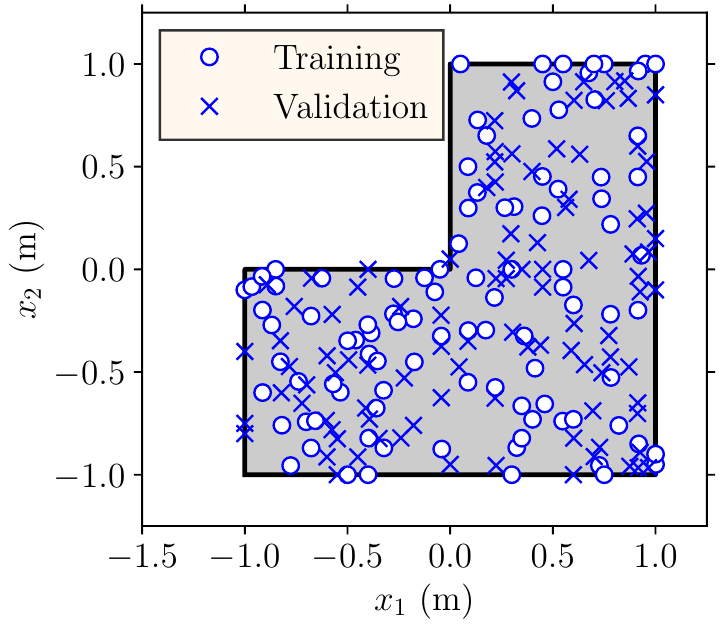}
    \caption{Measurement locations used during training and validation of DeepONets} 
    \label{fig:ExII_Meas_Loc1} 
        \end{subfigure} ~ 
    \centering 
    \begin{subfigure}[t]{0.45\textwidth}
        \centering 
        \includegraphics[scale=1]{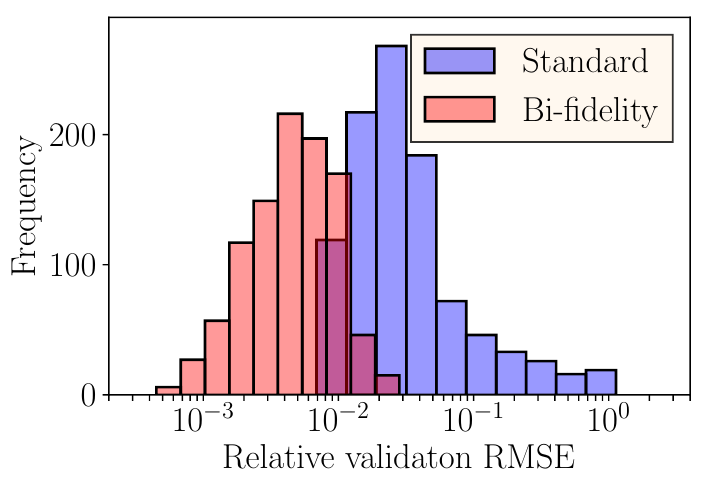}
     \caption{Comparison of histograms of relative validation error}
    \label{fig:ExII_hist1}



    \end{subfigure}%
    \caption{The training and validation locations and comparison of histograms of relative validation error for the uncertain thermal conductivity in Example 2. }
    \label{fig:ExII_hist_app1}
\end{figure}

In our first application, we model the thermal conductivity as 
\begin{equation} 
    k(\xm) = k_0 + \exp(z(\xm)), 
\end{equation} 
where $k_0=400$ $\text{W(m}\cdot\text{K)}^{-1}$ and $z(\xm)$ is a zero mean Gaussian random field with a square exponential covariance function $K(\xm,\xm^\prime)$ 
\begin{equation}
    K(\xm,\xm^\prime) = \exp\left(-\frac{1}{2} \sum_{i=1}^2 \left( \frac{x_i-x_i^\prime}{l_i} \right)^2\right), 
\end{equation} 
where $l_i,~i=1,2,$ are length scale parameters, which we choose as $0.5$ for both dimensions. A Karhunen-Lo{\`e}ve (KL) expansion truncated at the 25th term that captures 97\% of the energy (defined as the sum of all eigenvalues) is used to generate realizations of the random field $z(\xm)$. For the low-fidelity model, we consider  
\begin{equation}
    -h_tk_0\nabla^2 T_l(\xm) + 2h_cT_l(\xm) + 2\epsilon\sigma T_l^4(\xm) = 2h_cT_a + 2\epsilon\sigma T_a^4 + Q_i^\mathrm{nom}(\xm), 
\end{equation} 
where $T_l$ is the temperature given by the low-fidelity model; and $Q_i^\mathrm{nom}(\xm)=Q_0\exp\left(-\sum_{i=1}^2x_i^2\right)$. Hence, the low-fidelity model is deterministic and we only need to solve it once. 
Here, the input to the trunk network of the bi-fidelity DeepONet that models the discrepancy (see Figure \ref{fig:Uncertain_systems}) contains 25 standard normal random variables used in the KL expansion of $z(\xm)$, $\xi_Q$ in \eqref{eq:Q}, and the location $\xm$ for predicting the temperature. We use $\Ntr=30$ realizations of the uncertain variables in the training dataset $\Dtr$ and $\Nval=1000$ realizations for the validation dataset $\Dval$. 
For each of these realizations, we measure the temperature of the plate at 200 locations marked using circles and crosses in Figure \ref{fig:ExII_Meas_Loc1}. During training, we use the temperature at 100 of these locations marked with circles and validate the DeepONets by predicting the temperature at the remaining 100 locations marked with crosses. 
We compare the result to the standard use of DeepONet as discussed in Section \ref{sec:back}, where a sample of $Q_i(\xm)$ and the right hand side of \eqref{eq:ExII_gov_eqn}, measured at the training locations, marked with circles in Figure \ref{fig:ExII_Meas_Loc1}, are used as input to the branch network.

Using the bi-fidelity DeepONet, the validation error $\varepsilon_\mathrm{val}$, as defined in \eqref{eq:val_rmse}, is $6.72\times10^{-3}$. 
However, the validation error $\varepsilon_\mathrm{val}$ in the standard use of DeepONet is $8.70\times10^{-2}$. Figure \ref{fig:ExII_hist1} shows the histograms comparing the errors from these two approaches, denoted as `standard' and `bi-fidelity', respectively. 
The validation error $\varepsilon_\mathrm{val}$ and the histograms show that the proposed approach improves the accuracy of the predictions by almost an order of magnitude. This is primarily due to the fact that modeling of the temperature discrepancy can be performed relatively accurately using DeepONets in the presence of a small training dataset. 


\begin{figure}[!htb]
    \centering 
    \centering
    \begin{subfigure}[t]{0.45\textwidth}
        \centering
        \includegraphics[scale=1]{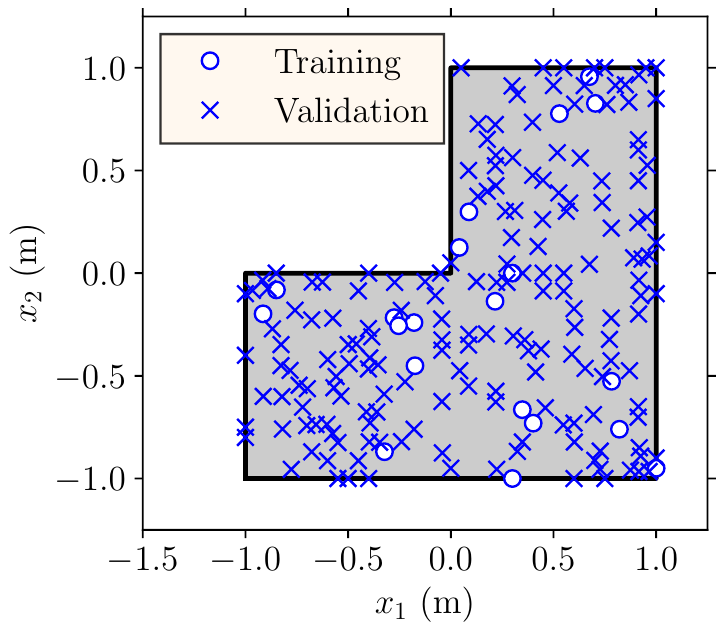}
    \caption{Measurement locations used during training and validation of DeepONets} 
    \label{fig:ExII_Meas_Loc2} 
        \end{subfigure} 
    ~ 
    \begin{subfigure}[t]{0.5\textwidth}
        \centering 
        \includegraphics[scale=1]{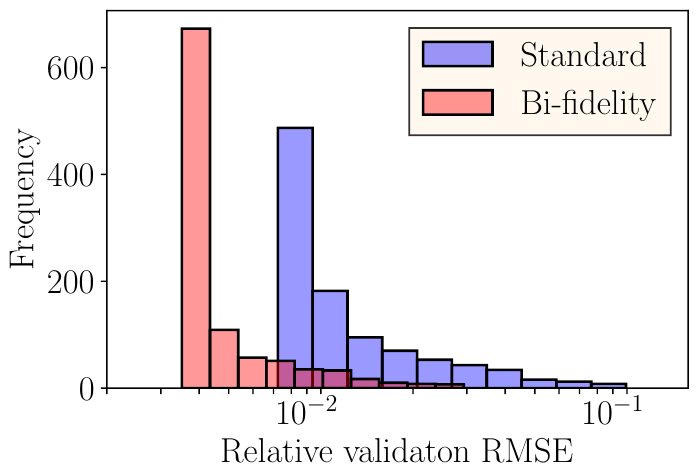}
        \caption{Application II: partially unknown system}
    \label{fig:ExII_hist2}

    \end{subfigure}%
    \caption{The training and validation locations and comparison of histograms of relative validation error for partially unknown system in Example 2. }
    \label{fig:ExII_hist_app2}
\end{figure}

\subsubsection{Application II: Partially Unknown System (Structural Uncertainty)} 

Next, we assume that the low-fidelity model is unaware of the radiation component in \eqref{eq:ExII_gov_eqn}, \textit{i.e.}, 
\begin{equation} 
    -h_tk_0\nabla^2 T_l(\xm) + 2h_cT_l(\xm) = 2h_cT_a + Q_i^\mathrm{nom}(\xm), 
\end{equation} 
where $Q_i^\mathrm{nom}(\xm)=Q_0\exp\left(-\sum_{i=1}^2x_i^2\right)$ is used. Hence, the low-fidelity model is deterministic and we only need to solve it once. Here, we use $\xm$ and $\xi_Q$ as the input to the trunk network. The training dataset $\Dtr$ and the validation dataset $\Dval$ consist of $\Ntr=3$ and $\Nval=1000$ data points, respectively, similar to the previous subsection. 
For each of these realizations, we measure the temperature of the plate at 200 locations marked using circles and crosses in Figure \ref{fig:ExII_Meas_Loc2}. During training, we use the temperature at 20 of these locations marked with circles and validate the DeepONets by predicting the temperature at the remaining 180 locations marked with crosses. 
When using DeepONet for modeling the discrepancy between the low-fidelity model and the true system, the validation error $\varepsilon_\mathrm{val}$ is $5.32\times10^{-3}$. 
The standard use of DeepONet, however, gives $\varepsilon_\mathrm{val}$ as $1.46\times10^{-2}$, one order of magnitude larger than the proposed bi-fidelity DeepONet. Figure \ref{fig:ExII_hist2} compares the histograms of the errors from these two approaches, illustrating the advantage of the bi-fidelity approach. 




\subsection{Example 3: Wind Farm} 

\begin{figure}[!htb]
    \centering
    \includegraphics[scale=1]{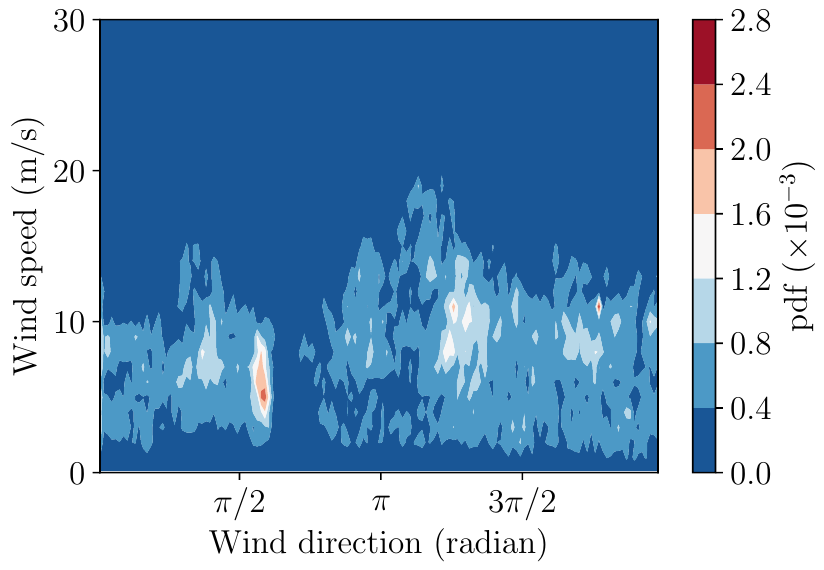}
    \caption{Joint probability density function (pdf) of the wind speed and direction measured at the Princess Amalia Wind Farm in the Netherlands used in Example 3.}
    \label{fig:wind_pdf} 
\end{figure}

In our third example, we model the power generated in a wind farm and consider uncertain wind speed, inflow direction, and yaw angle.  Estimating the expected power with respect to the joint distribution of wind speed and direction is a common calculation in determining the annual energy production of a wind plant; however, the number of function evaluations required often precludes the use of high-fidelity models in industry \cite{king2020probabilistic}.  Deliberately, yawing wind turbine rotors out of perpendicular to incoming wind is an emerging flow control technique called wake steering, which can improve the overall plant output by directing wind turbine wakes away from downwind turbines \cite{fleming2017field}.  The wake steering optimization problem is fraught with uncertainty due to misalignment in the nacelle-mounted wind vanes located just downwind of the rotor \cite{quick2020wake}. 
We consider two layouts of a wind farm with the number of turbines six and 64, respectively. For the probability distribution of uncertain wind speed and direction, we use 10 minute-average measurement data from the Princess Amalia wind farm in the Netherlands. The joint probability density function of the wind speed and direction estimated using these measurements is shown in Figure \ref{fig:wind_pdf}.

\begin{figure}[!htb]
    \centering
    \begin{subfigure}[t]{0.5\textwidth}
        \centering
\includegraphics[scale=1]{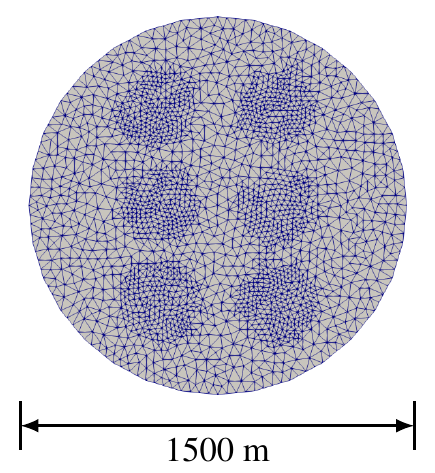}
        \caption{Two-dimensional mesh} 
        \label{fig:wind_2d_mesh} 
    \end{subfigure}%
    ~ 
    \begin{subfigure}[t]{0.5\textwidth}
        \centering
\includegraphics[scale=1]{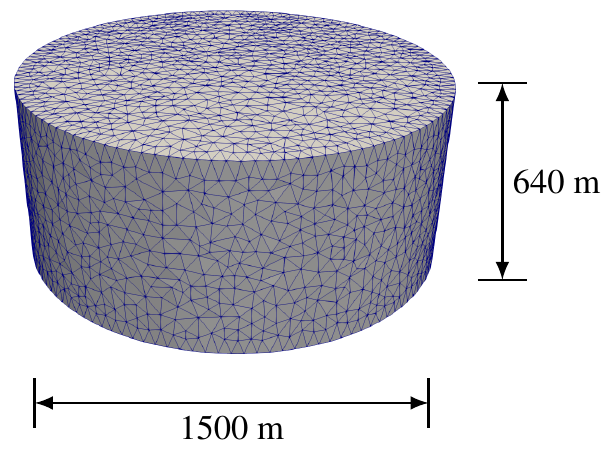}
        \caption{Three-dimensional mesh} 
        \label{fig:wind_3d_mesh} 
    \end{subfigure}
    \caption{Finite element meshes used in two- and three-dimensional simulations for Layout 1 with six turbines. }
    \label{fig:wind_mesh}
\end{figure}

For each realization of the wind speed, direction, and yaw angle, we use the {WindSE} toolkit \cite{king2017windse}, a Python package with FEniCS \cite{logg2012automated} as the backend solver, to generate the training and validation datasets. 
Two-dimensional (2D) Reynolds Averaged Navier-Stokes (RANS) simulations using the finite element mesh shown in Figure \ref{fig:wind_2d_mesh} are used to generate the low-fidelity data. Three-dimensional (3D) simulations using the mesh shown in Figure \ref{fig:wind_3d_mesh} are used to estimate the power generated from the true system. For the two- and three-dimensional simulations, a mixing length model \cite{prandtl1925bericht}\cite[7.1.7]{pope2000turbulent} is used for closing the turbulent stress term, with a uniform mixing length $l_m=15$~m. In the three-dimensional simulations, a velocity log-profile is used for the atmospheric boundary layer with Von-Karman constant $\kappa=0.4$. The log profile is such that the hub-height velocity is the value of the sample drawn from the pdf shown in Fig.~\ref{fig:wind_pdf}. In the two-dimensional simulations, the governing equations follow the 2D approximation of Boersma et al. \cite{boersma2018control} to approximate the flow field at hub-height, which is sufficient for estimating the power output.

We use a training $\Dtr$ and validation $\Dval$ datasets consisting of $\Ntr=100$ and $\Nval=1000$ realizations of the uncertain variables and power generated. 
The realizations of wind speed and wind direction are generated using the histograms from the dataset for the Princess Amalia wind farm in the Netherlands. 
We use a neural network with three hidden layers and 100 neurons each with ELU activation as the branch network and another neural network with two hidden layers and 100 neurons each with ELU activation as the trunk network in this example. In \eqref{eq:deep}, we truncate the summation at $p=60$ that produces the smallest validation error. Results for two layouts with six and 64 turbines are discussed next. 

\subsubsection{Layout 1: Six Turbines} 
\begin{figure}[!htb]
    \centering 
    \centering
    \begin{subfigure}[t]{0.5\textwidth}
        \centering
        \includegraphics[scale=1]{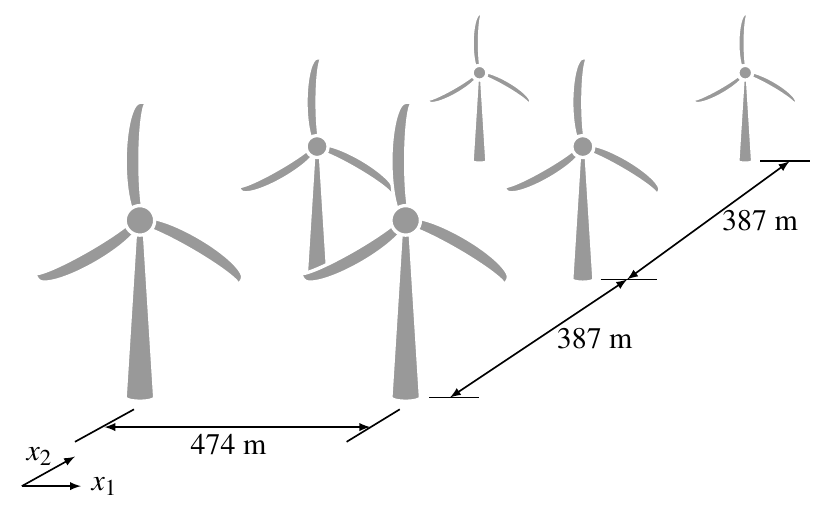}
  



    \caption{Wind farm layout}
    \label{fig:wind_farm}

    \end{subfigure}%
    ~ 
    \begin{subfigure}[t]{0.5\textwidth}
        \centering
  
  


\includegraphics[scale=1]{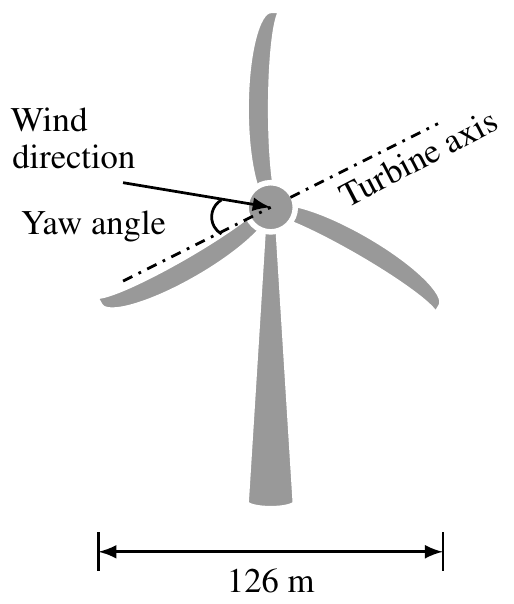}
    \caption{A single wind turbine}
    \label{fig:wind_turbine} 
    \end{subfigure}
    \caption{Schematic of the wind farm used in Example 3.} \label{fig:wind_farm_schem} 
\end{figure}

First, we model the power generated in a wind farm with six turbines arranged in a $3\times2$ grid as shown in Figure \ref{fig:wind_farm}. The wind turbines are placed 474 m apart in the $x_1$ direction and 387 m apart in the $x_2$ direction. We use the NREL 5 MW reference turbine \cite{jonkman2009definition}, which has a 126 m rotor diameter and 90 m hub height.
Figure \ref{fig:wind_turbine} shows a single turbine with \textit{yaw angle} between the wind direction and the turbine axis, which is used as an input parameter in this example.

We consider three cases here. In Case I, we only consider wind speed and direction as uncertain variables and assume yaw angle to be zero for all the turbines. In Case II, we assume the yaw angle to be uncertain and uniformly distributed between $[-\pi/2,\pi/2]$, but the same for all the turbines. In the final Case III, we assume each of the six turbines has uncertain yaw angles that are independent and identically distributed uniformly between $[-\pi/2,\pi/2]$. 

We compare the the results with a standard DeepONet that uses wind speed, direction, and yaw angle (for the last two cases) as branch network input and the turbine location as the trunk network input. Note that one 2D simulation of this layout takes 4.69 s (average of 1000 runs) compared to 3.54 min (average of 1000 runs) for one 3D simulation on the National Renewable Energy Laboratory's Eagle high-performance computing system (\href{https://www.nrel.gov/hpc/eagle-system.html}{https://www.nrel.gov/hpc/eagle-system.html}). 
Therefore, for a fair comparison, we train this DeepONet with a training dataset $\Dtr$ consisting of 103 realizations of the uncertain variables and corresponding power generated (\textit{i.e.}, with similar computational cost of generating the training dataset compared to the bi-fidelity training dataset used in the proposed approach). 

We list the results in Table \ref{tab:ExIII_results}, which shows that the proposed use of DeepONet to model the discrepancy between the power outputs from 2D and 3D simulations provides significant improvement in the validation error. Also, the standard use of DeepONet shows improvement when we add yaw angle as another random input. This is primarily due to the fact that yaw angle has a significant effect on the power generated, and the standard DeepONet is able to model that relation with increasing connection due to the addition of a new input node in the network. A similar but slight improvement in the proposed approach can also be noticed in Cases II and III in Table \ref{tab:ExIII_results}. 

\begin{sidewaystable}[!htb]
    \centering
    \begin{tabular}{ |c|c|c|c|c|c|c| } 
\hline
\Tstrut \multirow{2}{*}{Case} & \multirow{2}{*}{Method} & \multicolumn{2}{|c|}{Network input} & \multirow{2}{*}{$\varepsilon_\mathrm{val}$} & \multirow{2}{*}{Histograms of validation error} \Bstrut \\ \cline{3-4} 
\Tstrut & & Branch & Trunk & & \Bstrut \\
\hline
\Tstrut \multirow{12}{*}{I} & \multirow{6}{*}{Standard} & & & \multirow{6}{*}{$1.49\times10^{-1}$} & \multirow{12}{*}{ \includegraphics[scale=0.45]{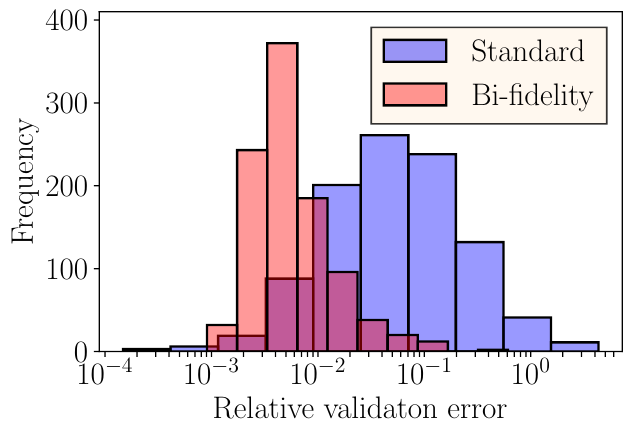}} \\ 
& & & & & \\ 
& & Wind speed, & Turbine & & \\
& & wind direction & location & & \\
& & & & & \\
& & & & & \Bstrut \\ \cline{2-5} 
\Tstrut & \multirow{6}{*}{Bi-fidelity} & & & \multirow{6}{*}{$1.03\times10^{-2}$} & \\
& & {Power} & Wind speed, & & \\
& & generated & wind direction, & & \\
& & {from} & turbine location & & \\
& & {2D simulations} & & & \\
& & & & & \Bstrut \\
\hline
\Tstrut \multirow{12}{*}{II} & \multirow{6}{*}{Standard} & & & \multirow{6}{*}{$4.98\times10^{-2}$} & \multirow{12}{*}{ \includegraphics[scale=0.45]{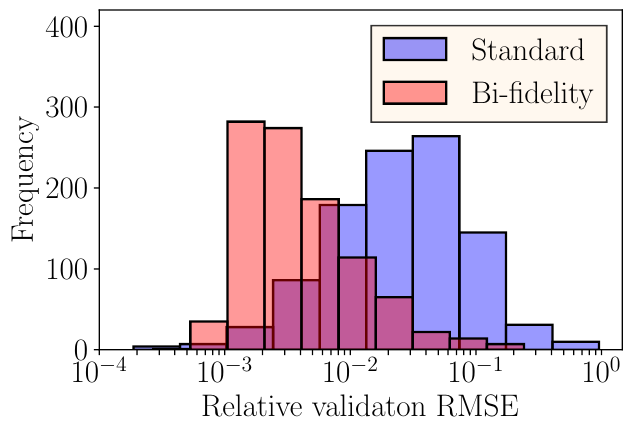}} \\ 
& & Wind speed, & & & \\ 
& & wind direction, & Turbine & & \\
& & yaw angle & location & & \\
& & & & & \\
& & & & & \Bstrut \\ \cline{2-5} 
\Tstrut & \multirow{6}{*}{Bi-fidelity} & & & \multirow{6}{*}{$8.55\times10^{-3}$} & \\
& & {Power} & Wind speed, & & \\
& & generated & wind direction, & & \\
& & {from} & yaw angle, & & \\
& & {2D simulations} & turbine location & & \\
& & & & & \Bstrut \\
\hline 
\Tstrut \multirow{12}{*}{III} & \multirow{6}{*}{Standard} & & & \multirow{6}{*}{$2.97\times10^{-2}$} & \multirow{12}{*}{\includegraphics[scale=0.45]{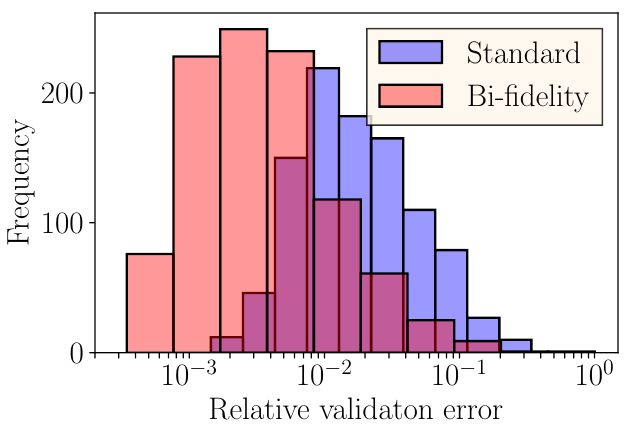}} \\ 
& & Wind speed, & & & \\ 
& & wind direction, & Turbine & & \\
& & individual & location & & \\
& & yaw angle & & & \\
& & & & & \Bstrut \\ \cline{2-5} 
\Tstrut & \multirow{6}{*}{Bi-fidelity} & & & \multirow{6}{*}{$9.06\times10^{-3}$} & \\
& & {Power} & Wind speed, & & \\
& & generated & wind direction, & & \\
& & {from} & individual yaw angle, & & \\
& & {2D simulations} & turbine location & & \\
& & & & & \Bstrut \\
\hline 
\end{tabular} 
\vspace{10pt} 
    \caption{Network inputs and validation results for three different cases studied in Example 3 for Layout 1 with six turbines. }
    \label{tab:ExIII_results}
\end{sidewaystable} 

\subsubsection{Layout 2: 64 Turbines} 
\begin{figure}[!htb]
    \centering
    \begin{subfigure}[t]{0.5\textwidth}
        \centering
        \includegraphics[scale=1]{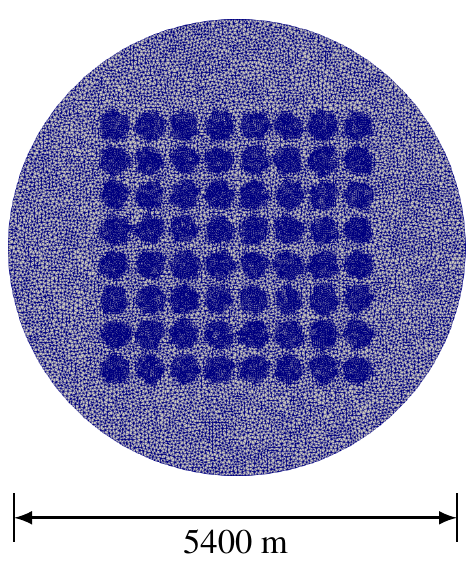}
        \caption{Two-dimensional mesh} 
        \label{fig:wind_2d_mesh_64} 
    \end{subfigure}%
    ~ 
    \begin{subfigure}[t]{0.5\textwidth}
        \centering
\includegraphics[scale=1]{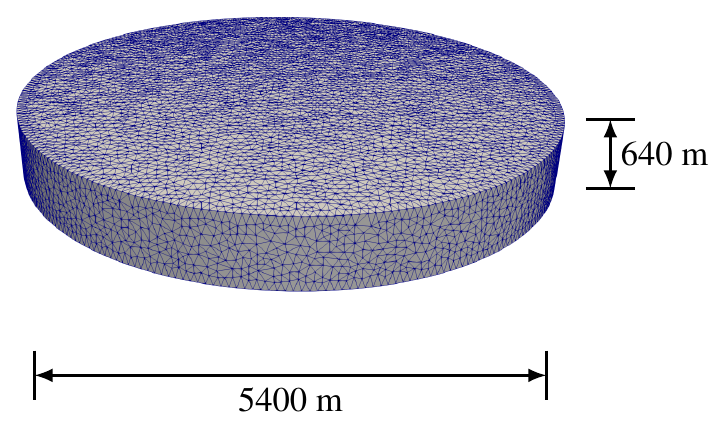}
        \caption{Three-dimensional mesh} 
        \label{fig:wind_3d_mesh_64} 
    \end{subfigure}
    \caption{Finite element meshes used in two- and three-dimensional simulations for Layout 2 with 64 turbines. }
    \label{fig:wind_mesh_64}
\end{figure}

\begin{figure}[!htb]
    \centering
    \includegraphics[scale=1]{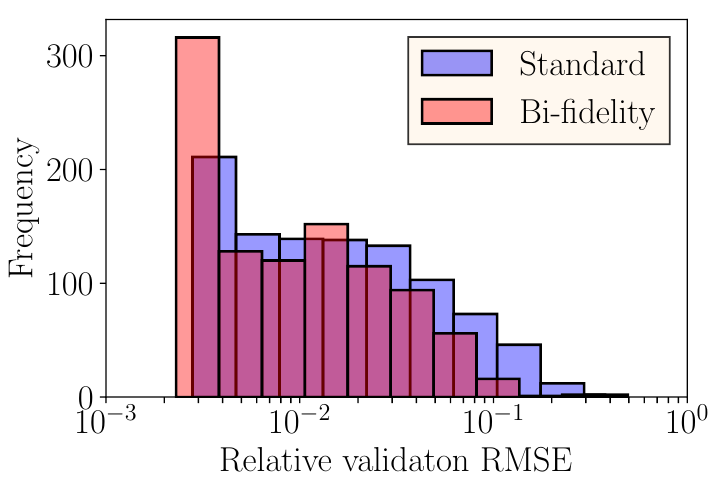}
    \caption{Histograms of the relative validation error obtained using the proposed use of DeepONet to model the discrepancy and the standard use of DeepONet in Example 3 with Layout 2 and 64 turbines.}
    \label{fig:wind_64_hist}
\end{figure}

In this layout, we consider 64 turbines in total arranged in a $8\times 8$ grid that is more characteristic of the size of a utility-scale wind plant. Other specifications of the turbine layout and the input parameters are the same as before. Figure \ref{fig:wind_mesh_64} shows the two- and three-dimensional meshes used to generate the training and validation datasets. 
The results from the proposed bi-fidelity training of DeepONets are compared with a DeepONet (denoted as `standard') trained only using a high-fidelity dataset with wind speed, direction, and yaw angle as branch network inputs and the turbine location as the trunk network input. 
Note that one 2D simulation of this layout takes 3.68 min. (average of 300 runs) compared to 1.15 hrs. (average of 300 runs) for one 3D simulation on the National Renewable Energy Laboratory's Eagle high-performance computing system. 
As before, for a fair comparison, we train the standard DeepONet with a training dataset $\Dtr$ consisting of 106 realizations of the uncertain variables and generated power to keep the cost of generating the training dataset similar to that of the bi-fidelity DeepONet. 
In this layout, the validation error $\varepsilon_\mathrm{val}$ using the proposed approach is $1.68\times10^{-2}$, which is about half of $2.97\times10^{-2}$, $\varepsilon_\mathrm{val}$ obtained using the standard DeepONet. 
Figure \ref{fig:wind_64_hist} shows the histograms of the validation errors. 

\FloatBarrier

\section{Conclusions} 

In this paper, we propose a bi-fidelity approach for modeling uncertain and partially unknown system responses using DeepONets. In this approach, we utilize a similar and computationally less expensive model of the physical system as a \textit{low-fidelity model}. The difference between the response from the \textit{low-fidelity model} and the true system is subsequently modeled using DeepONets. As is often the case, with low-fidelity model we can capture the important characteristics of the system's behavior. Therefore, we make a reasonable assumption in this study that the modeling of the discrepancy between the \textit{low-fidelity model} and the system's response can be done with a smaller high-fidelity dataset than a standard DeepONet to model the system's response in full to achieve similar validation error. In particular, we use DeepONets with low-fidelity model response, the locations, and the realizations of the uncertain variables as inputs. We apply the proposed approach to problems with parametric as well as with structural uncertainty, where some of the physics is unknown or not modeled. First, a nonlinear oscillator's response is modeled using this proposed approach, followed by the modeling of nonlinear heat transfer in a thin plate. Both examples show that the proposed approach provides an improvement in the validation error by almost an order of magnitude. Next, we model a wind farm with six wind turbines, where power generated from two-dimensional simulations is used as the \textit{low-fidelity} response and the true system's response is obtained using three-dimensional simulations. Again, the proposed approach provides significant improvement in validation error, showcasing the efficacy of the proposed approach. These three case studies demonstrate that the bi-fidelity DeepONet approach is a powerful tool in situations with limited high-fidelity data. 
In the future, we plan to develop a multi-fidelity extension of this approach to determine the effect of regularization in the loss function and modeling of complex multiphysics phenomena. 

\section*{Acknowledgments} 
This work was authored in part by the National Renewable Energy Laboratory, operated by Alliance for Sustainable Energy, LLC, for the U.S. Department of Energy (DOE) under Contract No. DE-AC36-08GO28308. This work was supported by funding from DOE's Advanced Scientific Computing Research (ASCR) program  under agreement DE-AC36-08GO28308. The research was performed using computational resources sponsored by the Department of Energy's Office of Energy Efficiency and Renewable Energy and located at the National Renewable Energy Laboratory. The views expressed in the article do not necessarily represent the views of the DOE or the U.S. Government. The U.S. Government retains and the publisher, by accepting the article for publication, acknowledges that the U.S. Government retains a nonexclusive, paid-up, irrevocable, worldwide license to publish or reproduce the published form of this work, or allow others to do so, for U.S. Government purposes. The work of AD was partially supported by the AFOSR grant FA9550-20-1-0138.

\appendix

\bibliographystyle{unsrt} 
\bibliography{references.bib}

\end{document}